\documentclass{article}

   \usepackage[nonatbib, final]{neurips_2025}

\usepackage[utf8]{inputenc} % allow utf-8 input
\usepackage[T1]{fontenc}    % use 8-bit T1 fonts
\usepackage{hyperref}       % hyperlinks
\usepackage{url}            % simple URL typesetting
\usepackage{booktabs}       % professional-quality tables
\usepackage{amsfonts}       % blackboard math symbols
\usepackage{nicefrac}       % compact symbols for 1/2, etc.
\usepackage{microtype}      % microtypography
\usepackage{xcolor}         % colors

\usepackage{graphicx}
\usepackage{amsmath}
\usepackage{graphicx}
\usepackage{geometry}                		     
\usepackage{setspace} % line spacing
\usepackage{mathtools}
\usepackage{amssymb}
\usepackage{bm,amsbsy}
\usepackage{algpseudocode, algorithm}

\usepackage[sort&compress,numbers]{natbib}

\renewcommand{\vec}[1]{\ensuremath{\pmb{#1}}}
\newcommand{\tr} {\ensuremath{^{\mathsf{T}}}}
\newcommand{\inv} {\ensuremath{^{-1}}}

\title{Separating the \emph{what} and \emph{how} of compositional computation to enable reuse and continual learning} % placeholder title

\author{%
  Haozhe Shan\thanks{Equal contribution.} \\
  Center for Theoretical Neuroscience \\
  Department of Computer Science\\
  Columbia University \\
  \texttt{hs3594@columbia.edu} \\
  \And
  Sun  Minni$^*$ \\
  Center for Theoretical Neuroscience \\
  Department of Neuroscience\\
  Columbia University \\
  \texttt{ms5724@columbia.edu} \\
  \And
  Lea Duncker \\
  Center for Theoretical Neuroscience \\
  Department of Neuroscience\\
  Columbia University \\
  \texttt{ld3149@columbia.edu} \\
}
\begin{document}

\maketitle

\begin{abstract}
The ability to continually learn, retain and deploy skills to accomplish goals is a key feature of intelligent and efficient behavior. 
However, the neural mechanisms facilitating the continual learning and flexible (re-)composition of skills remain elusive. 
Here, we study continual learning and the compositional reuse of learned computations in recurrent neural network (RNN) models using a novel two-system approach: one system that infers \emph{what} computation to perform, and one that implements \emph{how} to perform it. 
We focus on a set of compositional cognitive tasks commonly studied in neuroscience. 
To construct the \emph{what} system, we first show that a large family of tasks can be systematically described by a probabilistic generative model, where compositionality stems from a shared underlying vocabulary of discrete task epochs. 
%
% The shared epoch structure makes these tasks inherently compositional. 
%
% We first show that this compositionality can be systematically described by a probabilistic generative model. Furthermore, 
%
We develop an unsupervised online learning approach that can learn this model on a single-trial basis, building its vocabulary incrementally as it is exposed to new tasks, and inferring the latent epoch structure as a time-varying computational context within a trial. 
We implement the \emph{how} system as an RNN whose low-rank components are composed according to the context inferred by the \emph{what} system.
Contextual inference facilitates the creation, learning, and reuse of low-rank RNN components as new tasks are introduced sequentially, enabling continual learning without catastrophic forgetting. 
Using an example task set, we demonstrate the efficacy and competitive performance of this two-system learning framework, its potential for forward and backward transfer, as well as fast compositional generalization to unseen tasks. 
\end{abstract}

\section{Introduction}

% general intro on flexibility and compositionality
Humans and animals exhibit a remarkable ability to learn and retain new skills, and flexibly deploy them to accomplish goals in an ever-changing environment. 
A growing literature from neuroscience and human behavior suggests that mechanisms for contextual inference and task abstraction may play a crucial role for behavioral flexibility and learning \citep{heald2021contextual, wilson2014orbitofrontal, samborska2022complementary, niv2019learning, el2024cellular, flesch2023continual, zhou2021evolving, bein2025schemas, tian2025neural}. An abstract, task-relevant context could guide the selection and composition of different skills, while the maintenance of context-specific memories could counteract forgetting and aid continual learning \citep{heald2021contextual}.
Yet, the neural mechanisms facilitating the continual learning and flexible reuse of skills are not well understood. How do neural systems manage to expand their repertoire of skills without interference or forgetting, while maintaining the ability to access and compose them in new environmental contexts?
We hypothesize that multiple interacting learning systems with different objectives may contribute to overall learning and computation. Specifically, can inferences about the higher-level compositional structure of tasks be exploited for more robust and efficient learning?

We investigate the interplay of compositionality, contextual inference, and continual learning in recurrent neural networks (RNNs) using a two-systems approach: one that infers an abstract \emph{computational context} within a task family by parsing the compositional structure with probabilistic inference, and one that implements and retains the relevant computation for each context. This framework naturally maps onto a \emph{what} and \emph{how} architecture, previously explored in meta-learning \cite{he2019task}. The explicit separation into contexts (what) and computations (how) allows for flexible learning and compositional reuse. 
We focus our analyses on a family of cognitive and motor tasks commonly used in neuroscience \citep{yang2019task}. 
Here, we expand upon previous work in RNNs, where continual learning is still a major challenge \citep{cossu2021continual}, and compositional reuse of computational building blocks has only been shown to emerge implicitly over simultaneous training on multiple related tasks \cite{duncker2020organizing, driscoll2024flexible, riveland2024natural, costacurta2024structured}, rather than being explicitly implemented as a solution that facilitates sequential learning of tasks.

%context is typically taken to equal task identity, and compositional reuse of computational elements across tasks has been shown to emerge implicitly over training \cite{duncker2020organizing, driscoll2024flexible, riveland2024natural, costacurta2024structured}. Our framework, in contrast, makes the composition of different computational elements according to an inferred context explicit. We demonstrate that our approach facilitates continual learning when tasks are introduced sequentially over training and never revisited. In this setting, previous work has relied on selectively freezing weights \citep{driscoll2024flexible}, or slowing learning rates along particular dimensions in the system \citep{duncker2020organizing, kirkpatrick2017overcoming}.

% contributions & overview
The paper is organized as follows: 
We review related work and our contributions in section \ref{sec:background}. 
In section \ref{sec:task-model}, we formalize task compositionality through the development of a generative model for cognitive tasks. This allows us to define a computational context within a task family. We develop an unsupervised online learning and inference approach to directly infer context from the inputs and target responses that constitute a task, one trial at a time across a number of different but related tasks.
In section \ref{sec:full-model} we combine inference and learning in the task model with a contextually gated low-rank RNN. Training this architecture on a number of cognitive tasks, we demonstrate that our approach counteracts catastrophic forgetting and aids continual learning with signatures of both forward and backward transfer. We demonstrate competitive performance in comparison to continual learning approaches previously used in RNNs. We show that our architecture can rapidly learn to deploy known skills to solve novel tasks using only a few examples -- a property called compositional generalization \citep{wiedemer2023compositional, frankland2020concepts, dekker2022curriculum}. Finally, we discuss limitations and future directions in section \ref{sec:discussion}.

% more general impact/closing sentence 
Our work contributes to a more precise quantification of task compositionality, how it may shape computation, and how interactions across different learning systems can benefit continual learning and generalization. Ultimately, addressing these questions will help to elucidate flexibility and learning in biological systems.

\section{Related work and our contributions} 
\label{sec:background}
\textbf{Statistical theories of contextual inference.}
We follow an influential line of work on contextual inference in cognitive sciences \citep{gershman2010context, gershman2014statistical, heald2021contextual, heald2023contextual}. 
Contextual inference has been proposed as a model explaining many learning-related phenomena in classical conditioning \citep{gershman2010context} and motor control \cite{heald2021contextual}. 
Here, a latent discrete-valued random variable is used to describe distributional changes in task-relevant, continuous observations.
Given a set of observations, this variable can be inferred as a familiar or novel context to determine whether a learner may express and refine existing behaviors, or expand their repertoire using a new context.
% 
% by modeling context as a latent discrete-valued random variable to be inferred from continuous observations. The inferred context, in turn, controls whether the learner uses and refines one of the existing skills or creates a new one. 
% 
%
%
We extend this idea to the problem of learning to solve multiple common neuroscience tasks \cite{yang2019task, driscoll2024flexible}. To do this, we develop a formal description of these tasks in terms of a context variable:
% 
% a formalism of task compositionality and show how it captures some common neuroscience tasks.
Each task transitions through several contexts, each relevant for a different computation on a particular task epoch. 
A key aspect of our model is that these contexts can reappear in a compositional manner across different tasks, and may additionally share complex temporal dependencies within a given trial. 
%
% Furthermore, distinct from prior work in this lineage, continous observations for the current context explicitly depend on the expression of previous contexts within the same trial -- 
% 
%
In section \ref{sec:task-model} we show how this is crucial for capturing the distributional structure of neuroscience tasks, where, for example, the animal needs to compute a response based on information presented at an earlier time in the trial \citep{yang2019task, driscoll2024flexible}. %\lea{tried to edit a bit for clarity but not sure if it's cleaer now, or if the point can really be appreciated at this point?}

\textbf{Task-dependent modulation of neural networks.} The general notion of a separate system modulating computation in a neural network in a task-dependent manner has been previously studied. 
An example of this is the Task-Conditioned Hypernetwork \citep{von2019continual}, where the weights of a neural network are generated by a second, separate network based on modifiable task embeddings.
Other approaches often assume the network has its own learnable weights but that their involvement in computation is modulated, for instance via a gating mechanism \citep{tsuda2020modeling, masse2018alleviating, costacurta2024structured}. These works typically study tasks that can be solved by a single subcomponent (e.g., \citep{masse2018alleviating, flesch2023modelling, zheng2024rapid}) or the simple linear addition of outputs from multiple subcomponents (e.g., \citep{tsuda2020modeling, sandbrink2024flexible}). In our case, we study tasks that can only be solved via a composition of computations carried out by different subcomponents. 
%In the brain, the motor cortex has been suggested to be described as an RNN that generates different dynamical motifs under the control of a low-dimensional signal from the thalamus \citep{logiaco2021thalamic}. 
Our choice of architecture is similar to that of \citet{costacurta2024structured}, where an RNN receives a low-dimensional, time-varying `neuromodulatory' signal from a second RNN that adjusts how different low-rank components of its recurrent weights are combined. 
While \citet{costacurta2024structured} train this architecture end-to-end with backpropagation, we use the interpretable output of a contextual inference algorithm as the modulatory signal. 
In doing so, we make the compositional reuse of RNN components across contexts explicit, thereby facilitating continual learning and compositional generalization. Beyond RNNs, the `local module composition' scheme for feedforward networks \citep{ostapenko2021continual} is similar to our work in spirit, except that theirs requires passing each datum to all modules in order to determine the right composition to use and does not make use of an probabilistic inference procedure.

% We show that (1) instead of needing to learn this signal through backpropagation, we can use the interpretable output of a contextual inference algorithm; (2) building in compositionality explicitly allows continual learning and compositional generalization. 

\textbf{Continual learning in RNNs.} Continual learning is relatively understudied in RNNs and sequence-to-sequence problems more generally \citep{sodhani2020toward, cossu2021continual}. 
%
% Solutions to this problem
% Similarly to the feedforward case, it can be solved relatively simply by allowing the size of the network to grow as each task is learned \cite{tsuda2020modeling}. For comparisons to our work, 
In our comparisons, we focus on two previous continual learning approaches for networks with fixed dimensionality: 
Elastic weight consolidation (EWC) \cite{kirkpatrick2017overcoming}, which penalizes weight changes during future learning according to an estimate of how important each individual weight is for the performance of previous tasks. EWC is representative of a family of weight-regularization-based approaches \cite{ehret2020continual}. 
Another approach, Orthogonal Weight-Space Projection
(OWP), was developed in \cite{duncker2020organizing}, which pushes the network to solve different computations with activity in orthogonal subspaces. This method is representative of a family of approaches that modify weight updates during learning \citep{zeng2019continual, yu2020gradient,yang2023restricted}.
We found that our approach mitigates forgetting better than these methods, exhibits forward and backward transfer learning, and can rapidly learn new tasks by re-composing learned computations. 

% excluding methods that require growing the network as each task is learned \citep{tsuda2020modeling}. 
%
%\lea{might be repetitive with short review of methods in results section...maybe we can condense if we are pressed on space}

\section{Formalizing task compositionality via a probabilistic formulation} \label{sec:task-model}

\subsection{A generative model of compositional cognitive tasks}

% ------- FIGURE 1: TASK MODEL SET UP -------
\begin{figure}
\centering
\includegraphics[width=\linewidth]{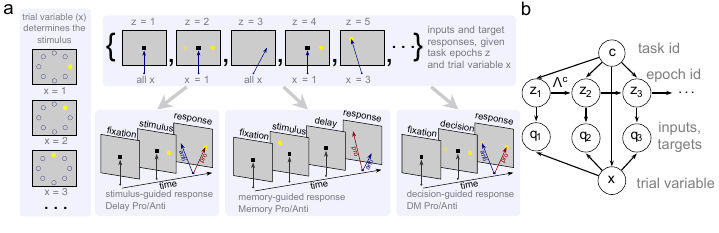}
 \vspace{-2ex}
\caption{A generative model of a family of cognitive tasks used in neuroscience. \textbf{a}: Schematic illustration of a set of cognitive tasks and their compositional nature. \textbf{b}: Directed Acyclic Graphical model (DAG) description of the generative model capturing the distribution over cognitive tasks we consider. The time-varying inputs and target responses of each task can be modeled as a mixture, where the observations for each task are composed of a set of discrete epochs, as illustrated in (\textbf{a}).}
\label{fig:task-model}
 \vspace{-3ex}
\end{figure}

Our goal is to study how the inference of a task-relevant computational context could guide the selection and composition of different skills in an RNN and aid continual learning. To address this question, we initially need to define what a task-relevant computational context is, and how it may be inferred as part of RNN training. 
We do this for a family of compositionally related cognitive tasks which have been popular both in experimental and computational neuroscience \cite{yang2019task, driscoll2024flexible, duncker2020organizing, costacurta2024structured, tafazoli2024building}. 

\begin{figure}[h]
\centering
\includegraphics{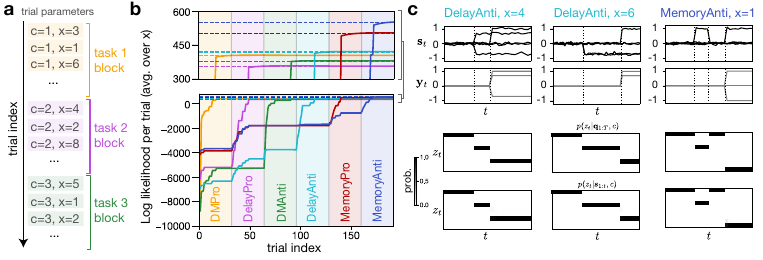}
\vspace{-1ex}
\caption{Online continual learning of the compositional structure of tasks. \textbf{a}: Schematics of online continual learning of a set of compositional tasks. \textbf{b}: Colored lines: log likelihood (LL) of tasks (averaged over trials) from the learned model over the course of learning.  Background colors: the task being trained at each trial. Dashed horizontal lines represent LLs computed using ground-truth parameters. \textbf{c} Single-trial epoch inference by the learned model. Top row: inputs and target responses of example trials. Dashed vertical lines indicate boundaries between epochs. Trial indices are dropped for brevity. Center row: output of training-time inference ($p(z_{t}|\vec{q}_{1:T},c)$). Bottom row: output of test-time inference ($p(z_{t}|\vec{s}_{1:t},c)$), which does not require access to the target responses or information from the future. Note that, since the $F$ and $M$ epochs in memory-guided tasks have indistinguishable observation models, they are merged by the learner.}
\label{fig:task-model-learning}
\vspace{-2ex}
\end{figure}

In these tasks, the learner receives a sequence of time-varying inputs $\vec{s}_{t=1,...,T}$ and needs to produce a target response sequence $\vec{y}_{t=1,...,T}$, following different distributions $p(\vec{s}_{1:T}, \vec{y}_{1:T}|c)$ depending on the task identity $c\in \{1, \dots, N_c\}$. These sequences are structured according to particular segments (often informally referred to as `epochs') such as a fixation period, a stimulus presentation period, or a response period. Each of these epochs defines particular distributions of inputs and target responses presented to the learner. What makes a family of tasks compositionally related is that they are composed from the same underlying set of epochs, albeit with task-dependent transitions from one epoch to the next. Fig. \ref{fig:task-model}a illustrates this schematically for the set of tasks we will consider throughout the paper. 
We first sought to formalize the notion of shared epochs (Fig. \ref{fig:task-model}b). We can describe complex dependencies in the distribution of inputs and target responses $p(\vec{s}_{1:T}, \vec{y}_{1:T}|c)$ for a task by introducing a discrete latent variable $z_t\in\mathbb{Z}^+$ denoting the task epoch at time $t$. We model epochs as evolving over time with task-dependent Markovian transitions $p(z_t=i | z_{t-1}=j, c) = \Lambda_{ij}^c$. Inputs and target responses $\vec{q}_t\equiv [\vec{s}_t,\vec{y}_t] \in \mathbb{R}^{D_q}$ are drawn from an epoch-dependent distribution. 
Each epoch may involve a number of different input values, reflecting e.g. different experimental conditions like reach direction or stimulus contrast. We therefore introduce dependence on an additional latent variable, the trial variable $x$, which can capture this trial-specific structure, with $\vec q_t \sim p(\vec q_t | z_t, x)$ (Fig \ref{fig:task-model}a). For example, $x$ may index a stimulus, which differs from one trial to the next and determines the distribution of inputs and target responses (see Figure \ref{fig:task-model-learning}c for examples). Importantly, sharing the same $x$ across the entire trial allows the model to capture dependencies between, e.g., the target responses in a later epoch and inputs from an earlier epoch. %We will return to this point below after introducing our example tasks.

Using this description of compositional cognitive tasks, a task-relevant context is naturally defined as the time-varying sequence of task epochs $z_t$. Thus, performing contextual inference requires learning the statistical structure and dependencies of the underlying generative model, and inferring the latent task epoch from observations $\vec q_t$. In the continual learning setting, this is essentially a problem of online learning and inference, which we address in section \ref{sec:online-learning}. We next give an overview of the specific tasks we consider throughout the rest of the paper. More detailed descriptions of the generative model and the specific task design are provided in Appendix \ref{appendix subsec: task generation model}.

\subsection{Overview of our task set}
\label{sec:task-details}
% \lea{minimal details here to follow results, know naming etc of tasks}

% brief intro about the tasks
We initially focus on six commonly studied neuroscience tasks \citep{yang2019task, duncker2020organizing, driscoll2024flexible, costacurta2024structured} that can be implemented in our generative framework. The tasks are schematized in Fig. \ref{fig:task-model}a and we will henceforth refer to them as \texttt{DelayPro}, \texttt{DelayAnti}, \texttt{MemoryPro}, \texttt{MemoryAnti}, \texttt{DMPro}, and \texttt{DMAnti}. 
All of these tasks require learning simple stimulus-response relationships, albeit with more complex temporal structure involving e.g. delayed responses. We summarize the different tasks in terms of their general goal and epoch structure in Table \ref{tab:tasks}. In practice, there is substantial inter-trial variability within a task due to variability in the amount of time spent in a given epoch, variability in the trial variable $x$, and additional variability in $\vec q_t=[\vec{s}_t,\vec{y}_t]$. 
All tasks involve a 5-dimensional time series of inputs $\vec s_t$ and a 3-dimensional time series of target responses $\vec y_t$, which take on noisy values around a piecewise constant mean. The value of the mean is dependent on the task epoch $z$ and trial variable $x$, chosen to reflect the underlying task goals. We refer to Appendix \ref{appendix subsec: task generation details} for a more detailed description of the tasks, their epoch structures, and distributional choices.

Note that inclusion of the trial variable $x$ has important consequences for how tasks are segmented into epochs. Considering the \texttt{MemoryPro/Anti} tasks: During the \textit{stimulus} epoch ($S$), the inputs reflect the trial variable $x$, which indexes the stimulus direction for each trial. During the `response period' ($R_{M,P/A}$), the target response is along/opposite the stimulus direction, thus making their dependencies on $x$ different. This necessitates treating them as two different epochs, $R_{M,P}$ and $R_{M,A}$ (see Table \ref{tab:tasks}). While this segmentation arises from our model of the complex statistical dependencies of cognitive tasks, different compositions may be possible using alternative formulations. For instance, one might imagine that all response-epochs could utilize the same context irrespective of how it relates to earlier stimuli, e.g. to move an effector in a particular direction. Indeed, empirical results suggest that this solution arises in RNN models trained on multiple tasks \citep{driscoll2024flexible}. %\lea{added this bit since maybe easier to make this point here than in discussion where it's currently missing?}

\begin{table}[]
    \centering
    \begin{tabular}{p{0.125\linewidth}|p{0.575\linewidth}|p{0.25\linewidth}}
       Task name  &  Description & Epoch sequence\\
       \hline
       \texttt{DelayPro}  & Respond towards stimulus direction $\theta$ after a delay. & $F \rightarrow S \rightarrow R_P$ \\ [0.5ex]
       \texttt{DelayAnti} & Same as \texttt{DelayPro} but respond towards $\theta + \pi$. & $F \rightarrow S \rightarrow R_A$ \\[0.5ex]
       \texttt{MemoryPro} & Memorize stimulus direction $\theta$ and respond towards it. & $F \rightarrow S \rightarrow M \rightarrow R_{M,P}$\\[0.5ex]
       \texttt{MemoryAnti} & Same as \texttt{MemoryPro} but respond towards $\theta + \pi$. & $F \rightarrow S \rightarrow M \rightarrow R_{M,A}$\\[0.5ex]
       \texttt{DMPro} & Compare amplitudes of two stimuli in directions $\theta$ and $\theta'$ and respond towards the direction of the higher amplitude. & $F \rightarrow S_{DM}  \rightarrow R_{DM,P}$\\[0.5ex]
       \texttt{DMAnti} & Same as \texttt{DMPro} but respond towards lower amplitude stimulus direction. & $F \rightarrow S_{DM}  \rightarrow R_{DM,A}$ \\ \hline
       
    \end{tabular}
    \hspace{1em}
    \caption{Our framework captures an ensemble of commonly used cognitive tasks with a shared compositional structure. The epochs are $F$: fixation; $S$: stimulus; $R_{P/A}$: response towards/opposite from the stimulus angle; $M$: memory; $R_{M,P/A}$: response towards/opposite from the memorized stimulus; $S_{DM}$: decision stimuli; $R_{DM,P/A}$: response towards/opposite the stronger stimulus.}
    \label{tab:tasks}
    % \vspace{-4ex}
\end{table}
% \vspace{2ex}

%In many cognitive tasks, the sequences of inputs and targets is typically composed according to different task epochs $z_t\in\mathbb{Z}^+$. Each task transitions through different epochs, each of which involves distinctive input and target statistics. A set of compositionally related tasks can share the same set of epochs but may have different transitions between them. Figure \ref{fig:task-model}a illustrates this schematically for a number of example tasks. We can translate this intuition into a probabilistic description of a task family in terms of a generative model. 
%
% necessary?

%

%Note that these tasks may be viewed as instances of sequence-to-sequence (seq2seq) tasks which have been of broad general interest in machine learning. \lea{add some refs?} In all tasks, the learner receives a sequence of time-varying inputs $\bm{s}_{t=1,...,T}$ and needs to produce a target output sequence $\bm{y}_{t=1,...,T}$. Different tasks (indexed by $c\in \mathbb{Z}^+$) can be distinguished by their distributions of inputs and targets $p(\bm{s}_{1:T}, \bm{y}_{1:T}|c)$. 
%%

% ------- FIGURE 2: TASK MODEL INFERENCE AND LEARNING -------

\subsection{\textit{What} system: online learning and inference of the compositional task structure} \label{sec:online-learning}

%\lea{unpack inference and learning approach for the task model}

We next consider how the compositional structure across tasks can be learned online (one trial at a time) and continually (one task at a time). %(see trials corresponding to one task before moving on to the next task). 
% The training data is a sequence of trials, $\{(\bm{q}^r_{1:T_r}, c_r,x_r)\}_{r=1,...,N_cN_{trials}}$, where $N_{trials}$ is the number of trials per task. 
The training data contains the sequences of inputs and target responses, as well as the task identity for each trial, $\{(\vec{q}^r_{1:T_r}, c_r)\}_{r=1,...,N_cN_{trials}}$, where $N_{trials}$ is the number of trials per task. We sometimes drop the trial index $r$ from notations for brevity. 
Given the training data, the goal of the \textit{what} system is to obtain estimates of the parameters that describe the distribution $p(\vec q_t | c)$ and transitions between epochs. It then infers the unobserved latent variables $z_t$ and $x$ on each trial. 

%
%$c_r=1$ for the first $N_{trials}$ trials, and $2$ for the second $N_{trials}$ trials and so on. 
% $x_r$ is drawn i.i.d. and uniformly from the 8 possible values. During the learning of each trial, the learner has access to $\bm{q}_{1:T_r}^r$, $c_r$ but not $x_r$. 

When training trials are available as batches across all trials and tasks, inference and learning can be solved using the classic Expectation-Maximization (EM) algorithm. EM performs a coordinate ascent on the complete data log-likelihood via alternating updates to the model parameters and to expected sufficient statistics of the latent variables \citep{dempster1977maximum}.
%We developed a learning and inference algorithm based on Expectation-Maximization (EM) and provide full details in Appendix \ref{appendix: learning algo}. 
To learn online from incoming streams of trials, we maintain estimates of the full-dataset sufficient statistics and update them incrementally after each trial. A pseudocode of the algorithm and details on the inference and parameter update equations are available in Appendix \ref{appendix subsec: contextual learning and inference}. To overcome convergence to bad local optima in the log-likelihood, we develop a structured incremental initialization procedure described in detail in Appendix \ref{appendix subsec: online learning initialization}.

We applied our algorithm to the problem of online continual learning of the 6 compositional tasks introduced in section \ref{sec:task-details}. To evaluate its performance, we computed the average single-trial log likelihood (LL) of different tasks, using the estimated parameters over the course of learning. We found that the model can recover LL on par with that from ground-truth parameters (Fig. \ref{fig:task-model-learning}b), and learning the structure of later tasks does not cause forgetting of previous tasks, even though the observation models of some epochs are shared across tasks. We note that this performance is robust to changes in task ordering (see examples in Appendix \ref{appendix: learning algo}), and the algorithm does not need to know the total number of tasks or epochs in advance, recognizing unfamiliar epochs on the fly. 
% \lea{should we say something about the model still learning after switching to a new task due to shared overall model/task family?}

The learned parameters can be used to infer the epoch identity $z_{t}$ during each trial, and we considered two types of inference. The more accurate inference (used to produce Fig. \ref{fig:task-model-learning}b) is $p(z_{t}|\vec{q}_{1:T},c)$. This uses both the inputs $\vec{s}_{1:T}$ and targets $\vec{y}_{1:T}$ for inference and produces a smoothing distribution in that the inference of $\vec{z}_t$ depends on data from $t'>t$. Since this requires access to the target responses, we call this `training time inference.' (Fig. \ref{fig:task-model-learning}c, center row). We also utilize a `test time inference' setting, where the learner has only access to the inputs up to the current time point, $\vec{s}_{1:t}$. This produces a filtering distribution, $p(z_{t}|\vec{s}_{1:t},c)$, which turns out to be highly accurate after learning (Fig. \ref{fig:task-model-learning}c, bottom row). In particular, it is able to disambiguate situations (e.g., $R_{M,P},R_{M,A}$ at the end of memory-guided response tasks have identical $\vec s_t$ statistics) by using the ground truth task label and the learned epoch transition structure for each task (e.g., knowing that \texttt{MemoryPro} and \texttt{MemoryAnti} tasks respectively end with $R_{M,P},R_{M,A}$). Hereafter, we will refer to this module that learns and infers underlying task structures as the `task model', and to the inferred epochs as `contexts'.

\section{Composing recurrent computation across multiple tasks via shared contexts} \label{sec:full-model}

%Our goal is to study how the inference of task-relevant context could guide the selection and composition of different skills in an RNN and aid continual learning. To address this question, we initially need to define what a task relevant context is, and how it may be inferred as part of RNN training. 

\begin{figure}
\centering
\includegraphics{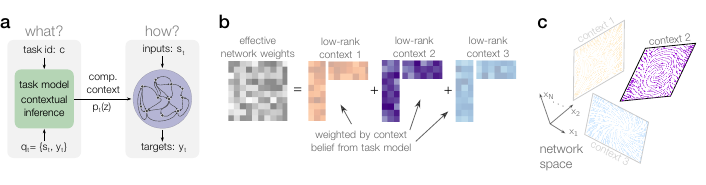}
 \vspace{-2ex}
\caption{Schematic illustration of two interacting learning systems and contextual gating of RNN dynamics. 
\textbf{a}: The task model (what) processes incoming input and target pairs $\vec q_t = \{\vec s_t, \vec y_t\}$ and infers the time-varying computational context of the current task. A downstream RNN (how) learns the relevant input-to-target transformation and is modulated by the task model's belief $p_t(z)$ over the current context. \textbf{b}: The effective recurrent weights of the downstream RNN are a linear combination of low-rank components, weighted by belief over each corresponding context. \textbf{c}: The gating by the task model causes the network to express different low-dimensional dynamics across different computational contexts, each of which can be re-expressed and reused to execute different tasks.}
\label{fig:what-how-setup}
\vspace{-2ex}
\end{figure}

The results presented so far were concerned with using information available based on the input and response pairs of the task to infer \emph{what} to do. This was achieved through the construction of a probabilistic model capturing the distribution across a family of compositionally related tasks, and the development of an algorithm that can infer a computational context within that task family. We now turn to the problem of using this context to guide the selection and composition of different computations in an RNN that learns \emph{how} to implement the relevant stimulus/response transformations. The general framework for how the two systems interact is illustrated in Fig. \ref{fig:what-how-setup}.

% ------- FIGURE 3: WHAT/HOW SET UP -------
\subsection{Recurrent neural network architecture}
We chose a simple RNN architecture, where the hidden-state activity $\vec{h}_t$ evolves in time according to 
\begin{equation}
\vec{h}_{t} = (1 - \alpha) \vec{h}_{t-1} + \alpha \left[W^{rec} \phi(\vec{h}_{t-1}) + W^{in} \vec{s}_t + \vec{b}^{in} + \sqrt{2 \alpha^{-1} \sigma_r^2}\ \vec{\xi}_t\right]
\label{eq:rnn_step}
\end{equation}
with decay rate $\alpha$, activation function $\phi$, external input $\vec{s}_t$, input bias $\vec{b}^{in}$ and uncorrelated Gaussian noise $\vec{\xi}_t$. The network activity is read out via
\begin{equation}
    \hat{\vec{y}}_t = W^{out} \phi(\vec{h}_t) + \vec{b}^{out},
    \label{eq:rnn_out}
\end{equation}
with output bias $\vec{b}^{out}$.
Given the external input and initial state, the network is trained to produce a target response time series $\vec{y}_t$ by minimizing the weighted mean squared error between $\vec{y}_t$ and $\hat{\vec{y}}_t$ (Appendix \ref{appendix subsec: loss and performance}). 

Inspired by related works on motif execution \cite{logiaco2021thalamic, costacurta2024structured}, the connectivity weights in our RNN are dynamically modulated by the posterior belief over computational contexts $p_t(z)$ inferred by the task model (Fig. \ref{fig:what-how-setup}a). More specifically, each computational context in our design corresponds to a set of weights: a low-rank component of recurrent weights $U_zV_z^T$, input weights $W_z^{in}$, input bias $\vec{b}_z^{in}$, output weights $W_z^{out}$, and output bias $\vec{b}_z^{out}$. The effective network weights at time $t$ are a weighted sum of these components (Fig. \ref{fig:what-how-setup}b): 
\begin{align}
    &W^{rec} = \sum_{z}p_t(z)\ U_z V_z^T \\
    &W^{in} = \sum_{z}p_t(z)\ W_z^{in}, \quad \vec{b}^{in} = \sum_{z}p_t(z)\ \vec{b}_z^{in}\\
    &W^{out} = \sum_{z}p_t(z)\ W_z^{out}, \quad \vec{b}^{out} = \sum_{z}p_t(z)\ \vec{b}_z^{out}
\end{align}
This architecture allows the network to express distinct dynamics for different computational contexts. (illustrated in Fig. \ref{fig:what-how-setup}c). And computations performed by this downstream \emph{how} system are selectively gated by the \emph{what} system.
Contrary to previous work, e.g. \citet{duncker2020organizing}, we do not constrain the RNN dynamics associated with one context to be non-interfering with those of another.

\subsection{Contextual modulation enables continual learning}

% ------- FIGURE 4: CONTINUAL LEARNING  -------
\begin{figure}
\centering
\includegraphics{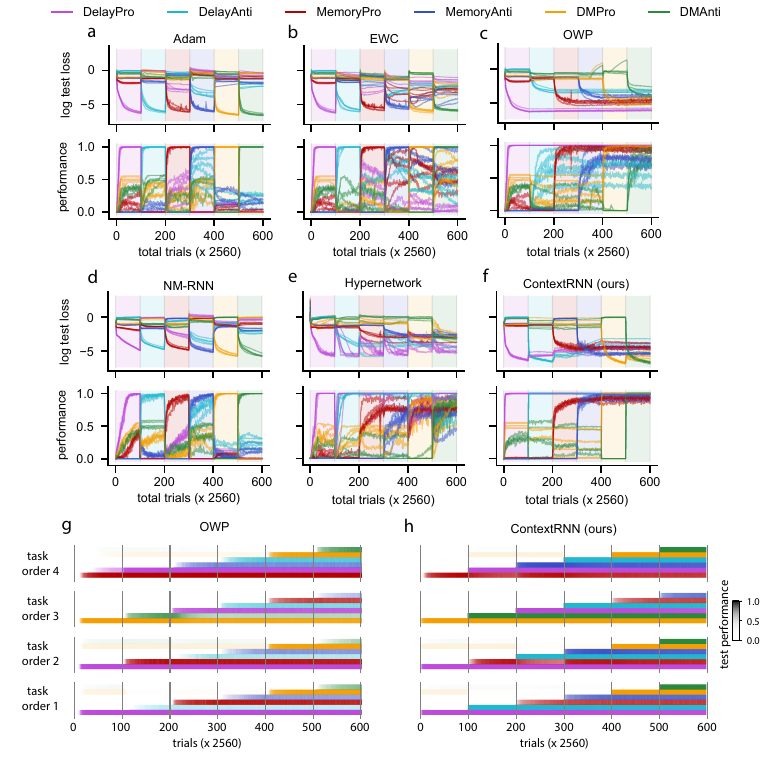}
 \vspace{-2ex}
\caption{Continual learning. \textbf{a-f}: log test loss (colored lines, first row) and test performance (colored lines, second row) throughout sequential training, each plot shows results for five random seeds. Background colors denote the task being trained at each trial. \textbf{a}: Adam optimizer on general RNNs. \textbf{b}: Elastic Weight Consolidation (EWC) on general RNNs. \textbf{c} Orthogonal Weight-Space Projection (OWP) on general RNNs. \textbf{d}: Adam optimizer on neuromodulated RNN. \textbf{e} Adam optimizer on Hypernetwork RNN. \textbf{f}: task model with context modulated RNNs (ours).  \textbf{g-h}: Color-coded test performance during sequential training of four different task orders. Each row color-codes the average test performance across five random seeds of a specific task over training. We compare OWP (\textbf{g}) with our method (\textbf{h}).} 
\label{fig:rnn-continual-learning}
\vspace{-2ex}
\end{figure}

Given the model architecture above, we now turn to the problem of training both systems in parallel on the set of tasks introduced in section \ref{sec:task-details}. As before, all training was performed sequentially, where $\{\vec s_t, \vec y_t\}$ pairs from only one task were shown to the network for a number of trials until the task identity switched and the previous tasks were never revisited (illustrated in Fig. \ref{fig:task-model-learning}a). 
During sequential training of each task, we first ran the incremental EM algorithm (Algorithm \ref{alg:cap}) for the task model over a whole batch, then computed contextual inference $p_t(z) := p(z_t|\vec{s}_{1:t}, \vec{y}_{1:t}, c)$ for the batch to instruct the downstream RNN. During testing, the RNN instead received test time inference $p_t(z) := p(z_t|\vec{s}_{1:t}, c)$ from the task model, which did not make reference to the target response $\vec y_t$. 

We compared the performance of our approach to five baselines and summarized the results in Fig. \ref{fig:rnn-continual-learning}, which shows the log loss and performance (definition see Appendix \ref{appendix subsec: loss and performance}) on test trials. 

There of the baselines operating on general RNNs governed by update equations (\ref{eq:rnn_step}) and (\ref{eq:rnn_out}), where $W^{rec}, W^{in}, \vec{b}^{in}, W^{out}, \vec{b}^{out}$ are unconstrained and shared across tasks. The general RNNs additionally receive a task identity input $W^c \vec{c}$, where $\vec{c}$ is a one-hot encoding of the current task. Under our default choices of hyperparameters (Appendix \ref{appendix subsec: RNN architecture}), these unconstrained RNNs have approximately twice as many trainable parameters as our context-modulated RNN.
The first baseline is to naively train general RNNs using the Adam optimizer \cite{kingma2014adam} without any additional measures for continual learning. In this case, interference across tasks led to catastrophic forgetting, as is readily visible from the log-loss and the task performance, which quickly degraded after a change in task identity (Fig. \ref{fig:rnn-continual-learning}a).  
The second baseline is Elastic Weight Consolidation (EWC) from \citet{kirkpatrick2017overcoming}, a continual learning approach which selectively slows down learning rates of single network weights deemed important on previous tasks. Due to the challenging nature of continual learning in RNNs, the simple importance-weighted approach failed to mitigate catastrophic forgetting (Fig. \ref{fig:rnn-continual-learning}b). 
For a baseline that specifically addressed continual learning in RNNs, we included the approach from \citet{duncker2020organizing}, which selectively slows learning rates in subspaces the network explored on previously learned tasks by projecting away these dimensions from the weight update throughout learning. We refer to this method as \emph{Orthogonal Weight-Space Projection} (OWP). While this approach outperformed Adam and EWC, the network suffered from capacity limitation where tasks progressively got harder to optimize. As a result, it became harder for OWP to achieve proficient task performance across all tasks (Fig. \ref{fig:rnn-continual-learning}c).

The other two baselines are more similar to our two-system approach in that they have a dedicated component that receives the task identity input and controls the weights used by an RNN that only receives the task input $\bm{s}_t$.  
One of them is neuromodulated RNNs from \citet{costacurta2024structured}, in which a neuromodulatory subnetwork RNN receives task identity input and outputs a signal that dynamically scales the low-rank recurrent weights of an output-generating RNN. Both subnetwork RNNs are jointly optimized end-to-end. While this design is similar to our two-system approach, it did not automatically solve continual learning and quickly forgot previously learned tasks when trained on new tasks (Fig. \ref{fig:rnn-continual-learning}d).
The fifth baseline uses hypernetworks for continual learning \citep{von2019continual, ehret2020continual}. A hypernetwork receives distinct learnable embeddings for each task and outputs the task-dependent weights for a target network (an output-generating RNN in our case). Changes to the weights for previously learned tasks are penalized when learning new tasks. This method showed limited improvement over EWC and forgot some earlier tasks at the end of training (Fig. \ref{fig:rnn-continual-learning}e).  
By contrast, Fig. \ref{fig:rnn-continual-learning}f shows results from our algorithm, which combines the task model with the context-modulated RNN. Selectively modulating the RNN based on the inferred context mitigated catastrophic forgetting and allowed the network to maintain high performance on all previously learned tasks throughout sequential training. This result was robust to different task training orders (Fig. \ref{fig:rnn-continual-learning}g, h).  

\subsection{Contextual inference facilitates forward \& backward transfer} % come up with a better name

% ------- FIGURE 5: TRANSFER LEARNING  -------
\begin{figure}
\centering
\includegraphics{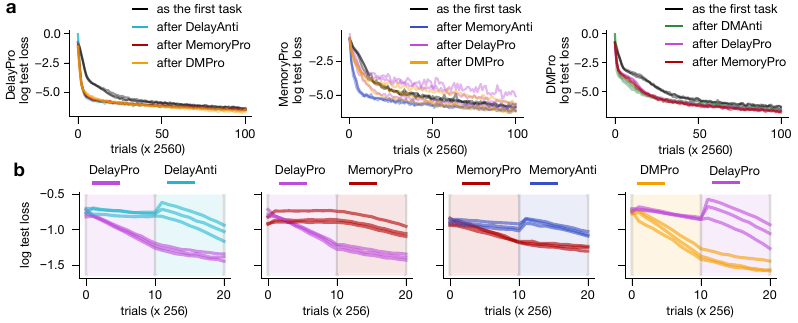}
 \vspace{-1ex}
\caption{Transfer learning. \textbf{a}: The log test loss during training of a task (specified in y-axis label), either as the first task (black) or after training another task (coded by color). We observe faster training after pre-training on tasks that share similar epochs, illustrating forward transfer.
 \textbf{b}: The log test loss during sequential training of two tasks, when each tasks is trained for fewer numbers of trials than our default choice. The loss of the previous tasks continues to decrease after switching to a new task, illustrating backward transfer. We plot results from three random seeds for each task order.}
\label{fig:transfer-learning}
\vspace{-3ex}

\end{figure}

We next asked whether, given the compositional structure of the task design, a shared computational context across tasks could facilitate transfer learning. Positive forward transfer refers to the phenomenon where the test loss for a given task decreases more rapidly when it is trained after another task, compared to when it is trained from scratch.
We observed positive forward transfer between nearly all pairs of tasks (Fig. \ref{fig:transfer-learning}a and Appendix \ref{appendix: suppl results}). An exception was \texttt{MemoryPro/Anti} tasks, which exhibited reliable forward transfer primarily when paired with another \texttt{Memory} task. When instead training after \texttt{Delay} or \texttt{DM} tasks, which share fewer overlapping epochs with \texttt{Memory} tasks, the model's test loss on \texttt{Memory} occasionally converged to a slightly higher final value after a rapid initial decrease relative to training from scratch. This phenomenon was also observed in \citet{duncker2020organizing}. 
When each task was trained for fewer trials (switching to a new task before proficiency was reached), we found that its test loss continued to decrease during training on the subsequent task (Fig. \ref{fig:transfer-learning}b and Appendix \ref{appendix: suppl results}), as long as overlapping computational contexts were revisited. This indicates positive backward transfer. Existing continual learning approaches rarely exhibit backward transfer \citep{lin2022beyond}. Indeed, repeating the same experiments using OWP failed to produce improvements on old tasks when learning new ones (see Appendix \ref{appendix: suppl results}).

\subsection{Contextual selection facilitates compositional generalization on new tasks}

\begin{figure}
\centering
 % \begin{minipage}[c]{0.6\textwidth}
    \includegraphics[width=0.75\linewidth]{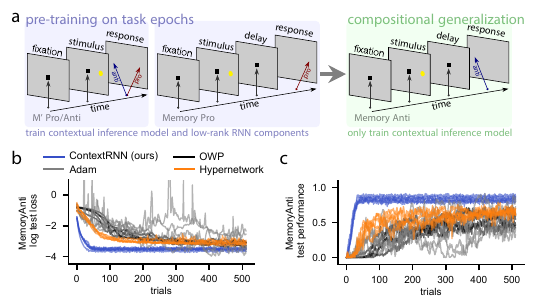}
  % \end{minipage}\hfill
  % \begin{minipage}[c]{0.35\textwidth}
  \vspace{-2ex} 
    \caption{Compositional generalization. \textbf{a}: Schematic of pre-training on a set of tasks and learning via contextual inference on a new task. \textbf{b-c}: The log loss (\textbf{b}) and task performance (\textbf{c}) on test trials of the new task as a function of trials used to train the task model (ContextRNN) compared to sequentially training the full models using OWP, Adam and Hypernetwork.}
    % \vspace{-2ex}
	\label{fig:few-shot}
  % \end{minipage}
  % \vspace{-3ex}
\end{figure}

As a final experiment, we evaluated the ability of our model to reuse previously learned contexts when encountering new tasks. The ability to do so is called \emph{compositional generalization}, and has been of major interest in neuroscience \cite{tafazoli2024building, riveland2024natural, driscoll2024flexible} and machine learning more generally \citep{wiedemer2023compositional, frankland2020concepts, dekker2022curriculum}.
%
% To evaluate compositional generalization in our model, we test its ability to reuse previously learned contexts when encountering new tasks. 
%
Specifically, we introduced a variant of the memory-guided response task in which the memory epoch ($M$) is omitted. These variants are denoted as \texttt{M'Pro} ($F \rightarrow S \rightarrow R_{M,P}$) and \texttt{M'Anti} ($F \rightarrow S \rightarrow R_{M,A}$) (Figure \ref{fig:few-shot}a; note that they are different from \texttt{DelayPro}/\texttt{DelayAnti} tasks in that the stimulus is not shown during response epochs). We initially trained the model sequentially on \texttt{M'Pro}, \texttt{M'Anti}, and \texttt{MemoryPro} ($F \rightarrow S \rightarrow M \rightarrow R_{M,P}$), and then assessed whether it can rapidly learn a new task, \texttt{MemoryAnti} ($F \rightarrow S \rightarrow M \rightarrow R_{M,A}$), within only a few trials. During this final stage, we froze the downstream RNN and updated only the task model, thereby testing whether the model could compose previously learned contexts to solve the novel task. Note that in principle, the freezing of RNN weights was not necessary and that the number of trials needed for learning the task model was generally much smaller than a typical batch size used for training the RNN. % \lea{fact check me?} \minni{batch size is 256, but here I trained the task model for 512 trials in fig 6b/c. It did converge within 40 trials.}
Our method reached approximately 83\% accuracy within as little as 40 trials on average. In contrast, baselines (also pretrained on the \texttt{M'Pro/Anti} and \texttt{MemoryPro} tasks) using Adam, OWP and Hypernetworks respectively achieved only 56\%, 53\%, 64\% accuracy after 512 trials of full model training (Figure \ref{fig:few-shot}b, c). These results demonstrate that our model can rapidly generalize to new tasks by recombining previously learned components.

% ------- FIGURE 6: FEW SHOT learning -------

\section{Discussion}
\label{sec:discussion}
How does the brain learn to solve tasks through compositional computation? We approached this question by first developing a formalism of compositional tasks that is sufficiently expressive to capture a suite of common neuroscience tasks as compositions from a shared vocabulary of task epochs. We then showed how it is possible to learn and solve these tasks continually using a dual-system approach: 
First, a \emph{what} system performs online probabilistic inference of the relevant compositional structure of tasks -- a time-varying, low-dimensional computational context corresponding to the task epoch.
This signal is then used to create or (re-)select computational components (one for each context) in a second \emph{how} system.
%
% Then, a low-rank RNN uses this signal to create, re-activate, or modifycomputational components, one for each epoch. 
%
Numerical experiments suggest that incorporating compositionality and contextual inference in this way allows the network to solve the challenging problem of continual learning of compositional tasks. Our approach can utilize knowledge transfer to future and previously encountered tasks, and exhibits potential for compositional generalization.

% trying to add a sentence that will link this back to neuro relevant to preface results
Studying interactions across different systems and how this may benefit learning, computation, and compositional generalization is also of substantial interest in neuroscience  \cite{tafazoli2024building, tian2025neural, sun2023organizing} . 
%
%Our work provides a model of two separate systems, each with different learning objectives, whose interaction during learning and task execution produces advantages for continual learning and compositional generalization.
%
While we don't directly map our approach to particular candidate brain areas, our work relates to an extensive neuroscience literature in this space. 
% Laureline work
Previous theoretical work has proposed that contextual gating of pattern-generating dynamics for motor sequences can aid flexible recomposition and learning, and be implemented via thalamocortical loops modulated via selective inhibition from basal ganglia \citep{logiaco2021thalamic}. A similar architecture would be possible for our model to extend these ideas from the motor domain to flexible cognitive tasks (see also \cite{zheng2024rapid}), and determine how the inhibitory signal from basal ganglia may arise according to contextual information from the \emph{what} system. More generally, there is a large literature on action selection, implicating basal ganglia circuits in the selection and composition of motor sequences \citep{sharpe2019integrated,markowitz2023spontaneous}. Other lines of work have studied interactions between hippocampus and prefrontal circuits for learning and generalization \citep{samborska2022complementary, whittington2020tolman, whittington2025tale}, and how abstract task states in prefrontal- \citep{el2024cellular, bein2025schemas} or orbitofrontal cortex \citep{niv2019learning, wilson2014orbitofrontal,zhou2021evolving} may be used to guide goal-directed behavior, computation, and learning.

% SMA M1
% BG Thal/Cortex
% PFC hippocampus

%%
%Evidence for separate \emph{what} and \emph{how} systems in the brain has been reported in the motor system \cite{russo2020neural}, where the supplementary motor area may plan a sequence of motor trajectories, each of which is more directly controlled by the primary motor cortex.
%%
%Other brain regions that may play the role of a \emph{what} system include the striatum \cite{sharpe2019integrated}, long known to be important for action selection, and the orbitofrontal cortex, thought to send information about task identity to other areas \cite{niv2019learning}. 
%%
%The specific mechanism by which our \emph{what} system modulates the \emph{how} system, gating of low-rank weight components, has been proposed as a model of how the striatum controls cortical dynamics via thalamocortical loops \cite{logiaco2021thalamic} and may be a general form of neuromodulation \cite{costacurta2024structured}. 
%%
%Our work provides a concrete model of how such a dual-system arrangement solves common neuroscience tasks as well as its advantages in terms of continual learning and compositional generalization. It also demonstrates how compositional computation in the brain \cite{tafazoli2024building, tian2025neural} can arise. \lea{still planning on editing this a bit}

\textbf{Limitations and future work.} 
A limitation of our work is that we can only infer computational context correctly when this is observable from the observed input and target response pairs of a task. When different epochs map onto the same observations for all $x$ (as is the case in our setting for e.g. fixation and memory epochs), the task model cannot infer that they are different. A possible future extension to overcome this would be to incorporate feedback from the RNN to the task model. A large error produced by the RNN for a familiar context could provide valuable information when different contexts are difficult to distinguish based on external observations alone. %Incorporating such information in the contextual inference algorithm is an important future direction. 
Similarly, the segmentation of tasks into epochs is not necessarily unique and depends on our modeling approach. A different notion of shared compositional structure resulting in a different task model may result in a different segmentation and opportunities for reuse.
In addition, we have assumed the trial variable $x$ to be a discrete variable, which simplifies the analysis by allowing us to express observation models for different $z,x$ as a look-up table of means. Assuming continuous $x$ may allow the model to capture richer tasks and improve scalability under complex stimulus distributions.
Finally, it may be of value to model the inference of unfamiliar epochs in a more principled way by using non-parametric methods (e.g., \citep{gershman2010context, heald2021contextual}). A key challenge is to extend existing algorithms to our case where epoch emissions are controlled by a latent factor $x$ that needs to be consistent across epochs, and context variables are shared across multiple tasks.

\begin{ack}

We would like to thank Dan O'Shea and Julia Costacurta for feedback on the manuscript and helpful discussions. This work was supported by the National Science Foundation and by DoD OUSD (R\&E) under Cooperative Agreement PHY-2229929 (The NSF AI Institute for Artificial and Natural Intelligence), the Gatsby Charitable Foundation (GAT3708), the Kavli Foundation, and the Simons Foundation Collaboration on the Global Brain.

% Use unnumbered first level headings for the acknowledgments. All acknowledgments

% Do {\bf not} include this section in the anonymized submission, only in the final paper. You can use the \texttt{ack} environment provided in the style file to automatically hide this section in the anonymized submission.
\end{ack}

% \newpage
% \bibliographystyle{unsrt}
\bibliography{ref}

\begin{thebibliography}{48}
\providecommand{\natexlab}[1]{#1}
\providecommand{\url}[1]{\texttt{#1}}
\expandafter\ifx\csname urlstyle\endcsname\relax
  \providecommand{\doi}[1]{doi: #1}\else
  \providecommand{\doi}{doi: \begingroup \urlstyle{rm}\Url}\fi

\bibitem[Heald et~al.(2021)Heald, Lengyel, and Wolpert]{heald2021contextual}
James~B Heald, M{\'a}t{\'e} Lengyel, and Daniel~M Wolpert.
\newblock Contextual inference underlies the learning of sensorimotor
  repertoires.
\newblock \emph{Nature}, 600\penalty0 (7889):\penalty0 489--493, 2021.

\bibitem[Wilson et~al.(2014)Wilson, Takahashi, Schoenbaum, and
  Niv]{wilson2014orbitofrontal}
Robert~C Wilson, Yuji~K Takahashi, Geoffrey Schoenbaum, and Yael Niv.
\newblock Orbitofrontal cortex as a cognitive map of task space.
\newblock \emph{Neuron}, 81\penalty0 (2):\penalty0 267--279, 2014.

\bibitem[Samborska et~al.(2022)Samborska, Butler, Walton, Behrens, and
  Akam]{samborska2022complementary}
Veronika Samborska, James~L Butler, Mark~E Walton, Timothy~EJ Behrens, and
  Thomas Akam.
\newblock Complementary task representations in hippocampus and prefrontal
  cortex for generalizing the structure of problems.
\newblock \emph{Nature Neuroscience}, 25\penalty0 (10):\penalty0 1314--1326,
  2022.

\bibitem[Niv(2019)]{niv2019learning}
Yael Niv.
\newblock Learning task-state representations.
\newblock \emph{Nature neuroscience}, 22\penalty0 (10):\penalty0 1544--1553,
  2019.

\bibitem[El-Gaby et~al.(2024)El-Gaby, Harris, Whittington, Dorrell, Bhomick,
  Walton, Akam, and Behrens]{el2024cellular}
Mohamady El-Gaby, Adam~Loyd Harris, James~CR Whittington, William Dorrell, Arya
  Bhomick, Mark~E Walton, Thomas Akam, and Timothy~EJ Behrens.
\newblock A cellular basis for mapping behavioural structure.
\newblock \emph{Nature}, pages 1--10, 2024.

\bibitem[Flesch et~al.(2023{\natexlab{a}})Flesch, Saxe, and
  Summerfield]{flesch2023continual}
Timo Flesch, Andrew Saxe, and Christopher Summerfield.
\newblock Continual task learning in natural and artificial agents.
\newblock \emph{Trends in neurosciences}, 46\penalty0 (3):\penalty0 199--210,
  2023{\natexlab{a}}.

\bibitem[Zhou et~al.(2021)Zhou, Jia, Montesinos-Cartagena, Gardner, Zong, and
  Schoenbaum]{zhou2021evolving}
Jingfeng Zhou, Chunying Jia, Marlian Montesinos-Cartagena, Matthew~PH Gardner,
  Wenhui Zong, and Geoffrey Schoenbaum.
\newblock Evolving schema representations in orbitofrontal ensembles during
  learning.
\newblock \emph{Nature}, 590\penalty0 (7847):\penalty0 606--611, 2021.

\bibitem[Bein and Niv(2025)]{bein2025schemas}
Oded Bein and Yael Niv.
\newblock Schemas, reinforcement learning and the medial prefrontal cortex.
\newblock \emph{Nature Reviews Neuroscience}, pages 1--17, 2025.

\bibitem[Tian et~al.(2025)Tian, Garz{\'o}n, Rouse, Eldridge, Schieber, Wang,
  Tenenbaum, and Freiwald]{tian2025neural}
Lucas~Y Tian, Kedar~U Garz{\'o}n, Adam~G Rouse, Mark~AG Eldridge, Marc~H
  Schieber, Xiao-Jing Wang, Joshua~B Tenenbaum, and Winrich~A Freiwald.
\newblock Neural representation of action symbols in primate frontal cortex.
\newblock \emph{bioRxiv}, pages 2025--03, 2025.

\bibitem[He et~al.(2019)He, Sygnowski, Galashov, Rusu, Teh, and
  Pascanu]{he2019task}
Xu~He, Jakub Sygnowski, Alexandre Galashov, Andrei~A Rusu, Yee~Whye Teh, and
  Razvan Pascanu.
\newblock Task agnostic continual learning via meta learning.
\newblock \emph{arXiv preprint arXiv:1906.05201}, 2019.

\bibitem[Yang et~al.(2019)Yang, Joglekar, Song, Newsome, and
  Wang]{yang2019task}
Guangyu~Robert Yang, Madhura~R Joglekar, H~Francis Song, William~T Newsome, and
  Xiao-Jing Wang.
\newblock Task representations in neural networks trained to perform many
  cognitive tasks.
\newblock \emph{Nature neuroscience}, 22\penalty0 (2):\penalty0 297--306, 2019.

\bibitem[Cossu et~al.(2021)Cossu, Carta, Lomonaco, and
  Bacciu]{cossu2021continual}
Andrea Cossu, Antonio Carta, Vincenzo Lomonaco, and Davide Bacciu.
\newblock Continual learning for recurrent neural networks: an empirical
  evaluation.
\newblock \emph{Neural Networks}, 143:\penalty0 607--627, 2021.

\bibitem[Duncker et~al.(2020)Duncker, Driscoll, Shenoy, Sahani, and
  Sussillo]{duncker2020organizing}
Lea Duncker, Laura Driscoll, Krishna~V Shenoy, Maneesh Sahani, and David
  Sussillo.
\newblock Organizing recurrent network dynamics by task-computation to enable
  continual learning.
\newblock \emph{Advances in neural information processing systems},
  33:\penalty0 14387--14397, 2020.

\bibitem[Driscoll et~al.(2024)Driscoll, Shenoy, and
  Sussillo]{driscoll2024flexible}
Laura~N Driscoll, Krishna Shenoy, and David Sussillo.
\newblock Flexible multitask computation in recurrent networks utilizes shared
  dynamical motifs.
\newblock \emph{Nature Neuroscience}, 27\penalty0 (7):\penalty0 1349--1363,
  2024.

\bibitem[Riveland and Pouget(2024)]{riveland2024natural}
Reidar Riveland and Alexandre Pouget.
\newblock Natural language instructions induce compositional generalization in
  networks of neurons.
\newblock \emph{Nature Neuroscience}, 27\penalty0 (5):\penalty0 988--999, 2024.

\bibitem[Costacurta et~al.(2024)Costacurta, Bhandarkar, Zoltowski, and
  Linderman]{costacurta2024structured}
Julia Costacurta, Shaunak Bhandarkar, David Zoltowski, and Scott Linderman.
\newblock Structured flexibility in recurrent neural networks via
  neuromodulation.
\newblock \emph{Advances in Neural Information Processing Systems},
  37:\penalty0 1954--1972, 2024.

\bibitem[Wiedemer et~al.(2023)Wiedemer, Mayilvahanan, Bethge, and
  Brendel]{wiedemer2023compositional}
Thadd{\"a}us Wiedemer, Prasanna Mayilvahanan, Matthias Bethge, and Wieland
  Brendel.
\newblock Compositional generalization from first principles.
\newblock \emph{Advances in Neural Information Processing Systems},
  36:\penalty0 6941--6960, 2023.

\bibitem[Frankland and Greene(2020)]{frankland2020concepts}
Steven~M Frankland and Joshua~D Greene.
\newblock Concepts and compositionality: in search of the brain's language of
  thought.
\newblock \emph{Annual review of psychology}, 71\penalty0 (1):\penalty0
  273--303, 2020.

\bibitem[Dekker et~al.(2022)Dekker, Otto, and
  Summerfield]{dekker2022curriculum}
Ronald~B Dekker, Fabian Otto, and Christopher Summerfield.
\newblock Curriculum learning for human compositional generalization.
\newblock \emph{Proceedings of the National Academy of Sciences}, 119\penalty0
  (41):\penalty0 e2205582119, 2022.

\bibitem[Gershman et~al.(2010)Gershman, Blei, and Niv]{gershman2010context}
Samuel~J Gershman, David~M Blei, and Yael Niv.
\newblock Context, learning, and extinction.
\newblock \emph{Psychological review}, 117\penalty0 (1):\penalty0 197, 2010.

\bibitem[Gershman et~al.(2014)Gershman, Radulescu, Norman, and
  Niv]{gershman2014statistical}
Samuel~J Gershman, Angela Radulescu, Kenneth~A Norman, and Yael Niv.
\newblock Statistical computations underlying the dynamics of memory updating.
\newblock \emph{PLoS computational biology}, 10\penalty0 (11):\penalty0
  e1003939, 2014.

\bibitem[Heald et~al.(2023)Heald, Lengyel, and Wolpert]{heald2023contextual}
James~B Heald, M{\'a}t{\'e} Lengyel, and Daniel~M Wolpert.
\newblock Contextual inference in learning and memory.
\newblock \emph{Trends in cognitive sciences}, 27\penalty0 (1):\penalty0
  43--64, 2023.

\bibitem[Von~Oswald et~al.(2019)Von~Oswald, Henning, Grewe, and
  Sacramento]{von2019continual}
Johannes Von~Oswald, Christian Henning, Benjamin~F Grewe, and Jo{\~a}o
  Sacramento.
\newblock Continual learning with hypernetworks.
\newblock \emph{arXiv preprint arXiv:1906.00695}, 2019.

\bibitem[Tsuda et~al.(2020)Tsuda, Tye, Siegelmann, and
  Sejnowski]{tsuda2020modeling}
Ben Tsuda, Kay~M Tye, Hava~T Siegelmann, and Terrence~J Sejnowski.
\newblock A modeling framework for adaptive lifelong learning with transfer and
  savings through gating in the prefrontal cortex.
\newblock \emph{Proceedings of the National Academy of Sciences}, 117\penalty0
  (47):\penalty0 29872--29882, 2020.

\bibitem[Masse et~al.(2018)Masse, Grant, and Freedman]{masse2018alleviating}
Nicolas~Y Masse, Gregory~D Grant, and David~J Freedman.
\newblock Alleviating catastrophic forgetting using context-dependent gating
  and synaptic stabilization.
\newblock \emph{Proceedings of the National Academy of Sciences}, 115\penalty0
  (44):\penalty0 E10467--E10475, 2018.

\bibitem[Flesch et~al.(2023{\natexlab{b}})Flesch, Nagy, Saxe, and
  Summerfield]{flesch2023modelling}
Timo Flesch, David~G Nagy, Andrew Saxe, and Christopher Summerfield.
\newblock Modelling continual learning in humans with hebbian context gating
  and exponentially decaying task signals.
\newblock \emph{PLoS computational biology}, 19\penalty0 (1):\penalty0
  e1010808, 2023{\natexlab{b}}.

\bibitem[Zheng et~al.(2024)Zheng, Wu, Hummos, Yang, and
  Halassa]{zheng2024rapid}
Wei-Long Zheng, Zhongxuan Wu, Ali Hummos, Guangyu~Robert Yang, and Michael~M
  Halassa.
\newblock Rapid context inference in a thalamocortical model using recurrent
  neural networks.
\newblock \emph{Nature Communications}, 15\penalty0 (1):\penalty0 8275, 2024.

\bibitem[Sandbrink et~al.(2024)Sandbrink, Bauer, Proca, Saxe, Summerfield, and
  Hummos]{sandbrink2024flexible}
Kai Sandbrink, Jan Bauer, Alexandra Proca, Andrew Saxe, Christopher
  Summerfield, and Ali Hummos.
\newblock Flexible task abstractions emerge in linear networks with fast and
  bounded units.
\newblock \emph{Advances in Neural Information Processing Systems},
  37:\penalty0 6938--6978, 2024.

\bibitem[Ostapenko et~al.(2021)Ostapenko, Rodriguez, Caccia, and
  Charlin]{ostapenko2021continual}
Oleksiy Ostapenko, Pau Rodriguez, Massimo Caccia, and Laurent Charlin.
\newblock Continual learning via local module composition.
\newblock \emph{Advances in Neural Information Processing Systems},
  34:\penalty0 30298--30312, 2021.

\bibitem[Sodhani et~al.(2020)Sodhani, Chandar, and Bengio]{sodhani2020toward}
Shagun Sodhani, Sarath Chandar, and Yoshua Bengio.
\newblock Toward training recurrent neural networks for lifelong learning.
\newblock \emph{Neural computation}, 32\penalty0 (1):\penalty0 1--35, 2020.

\bibitem[Kirkpatrick et~al.(2017)Kirkpatrick, Pascanu, Rabinowitz, Veness,
  Desjardins, Rusu, Milan, Quan, Ramalho, Grabska-Barwinska,
  et~al.]{kirkpatrick2017overcoming}
James Kirkpatrick, Razvan Pascanu, Neil Rabinowitz, Joel Veness, Guillaume
  Desjardins, Andrei~A Rusu, Kieran Milan, John Quan, Tiago Ramalho, Agnieszka
  Grabska-Barwinska, et~al.
\newblock Overcoming catastrophic forgetting in neural networks.
\newblock \emph{Proceedings of the national academy of sciences}, 114\penalty0
  (13):\penalty0 3521--3526, 2017.

\bibitem[Ehret et~al.(2020)Ehret, Henning, Cervera, Meulemans, Von~Oswald, and
  Grewe]{ehret2020continual}
Benjamin Ehret, Christian Henning, Maria~R Cervera, Alexander Meulemans,
  Johannes Von~Oswald, and Benjamin~F Grewe.
\newblock Continual learning in recurrent neural networks.
\newblock \emph{arXiv preprint arXiv:2006.12109}, 2020.

\bibitem[Zeng et~al.(2019)Zeng, Chen, Cui, and Yu]{zeng2019continual}
Guanxiong Zeng, Yang Chen, Bo~Cui, and Shan Yu.
\newblock Continual learning of context-dependent processing in neural
  networks.
\newblock \emph{Nature Machine Intelligence}, 1\penalty0 (8):\penalty0
  364--372, 2019.

\bibitem[Yu et~al.(2020)Yu, Kumar, Gupta, Levine, Hausman, and
  Finn]{yu2020gradient}
Tianhe Yu, Saurabh Kumar, Abhishek Gupta, Sergey Levine, Karol Hausman, and
  Chelsea Finn.
\newblock Gradient surgery for multi-task learning.
\newblock \emph{Advances in neural information processing systems},
  33:\penalty0 5824--5836, 2020.

\bibitem[Yang et~al.(2023)Yang, Yang, Liu, Li, and Liu]{yang2023restricted}
Zeyuan Yang, Zonghan Yang, Yichen Liu, Peng Li, and Yang Liu.
\newblock Restricted orthogonal gradient projection for continual learning.
\newblock \emph{AI Open}, 4:\penalty0 98--110, 2023.

\bibitem[Tafazoli et~al.(2024)Tafazoli, Bouchacourt, Ardalan, Markov, Uchimura,
  Mattar, Daw, and Buschman]{tafazoli2024building}
Sina Tafazoli, Flora~M Bouchacourt, Adel Ardalan, Nikola~T Markov, Motoaki
  Uchimura, Marcelo~G Mattar, Nathaniel~D Daw, and Timothy~J Buschman.
\newblock Building compositional tasks with shared neural subspaces.
\newblock \emph{bioRxiv}, 2024.

\bibitem[Dempster et~al.(1977)Dempster, Laird, and Rubin]{dempster1977maximum}
Arthur~P Dempster, Nan~M Laird, and Donald~B Rubin.
\newblock Maximum likelihood from incomplete data via the em algorithm.
\newblock \emph{Journal of the royal statistical society: series B
  (methodological)}, 39\penalty0 (1):\penalty0 1--22, 1977.

\bibitem[Logiaco et~al.(2021)Logiaco, Abbott, and Escola]{logiaco2021thalamic}
Laureline Logiaco, LF~Abbott, and Sean Escola.
\newblock Thalamic control of cortical dynamics in a model of flexible motor
  sequencing.
\newblock \emph{Cell reports}, 35\penalty0 (9), 2021.

\bibitem[Kingma and Ba(2014)]{kingma2014adam}
Diederik~P Kingma and Jimmy Ba.
\newblock Adam: A method for stochastic optimization.
\newblock \emph{arXiv preprint arXiv:1412.6980}, 2014.

\bibitem[Lin et~al.(2022)Lin, Yang, Fan, and Zhang]{lin2022beyond}
Sen Lin, Li~Yang, Deliang Fan, and Junshan Zhang.
\newblock Beyond not-forgetting: Continual learning with backward knowledge
  transfer.
\newblock \emph{Advances in Neural Information Processing Systems},
  35:\penalty0 16165--16177, 2022.

\bibitem[Sun et~al.(2023)Sun, Advani, Spruston, Saxe, and
  Fitzgerald]{sun2023organizing}
Weinan Sun, Madhu Advani, Nelson Spruston, Andrew Saxe, and James~E Fitzgerald.
\newblock Organizing memories for generalization in complementary learning
  systems.
\newblock \emph{Nature neuroscience}, 26\penalty0 (8):\penalty0 1438--1448,
  2023.

\bibitem[Sharpe et~al.(2019)Sharpe, Stalnaker, Schuck, Killcross, Schoenbaum,
  and Niv]{sharpe2019integrated}
Melissa~J Sharpe, Thomas Stalnaker, Nicolas~W Schuck, Simon Killcross, Geoffrey
  Schoenbaum, and Yael Niv.
\newblock An integrated model of action selection: distinct modes of cortical
  control of striatal decision making.
\newblock \emph{Annual review of psychology}, 70\penalty0 (1):\penalty0 53--76,
  2019.

\bibitem[Markowitz et~al.(2023)Markowitz, Gillis, Jay, Wood, Harris,
  Cieszkowski, Scott, Brann, Koveal, Kula, et~al.]{markowitz2023spontaneous}
Jeffrey~E Markowitz, Winthrop~F Gillis, Maya Jay, Jeffrey Wood, Ryley~W Harris,
  Robert Cieszkowski, Rebecca Scott, David Brann, Dorothy Koveal, Tomasz Kula,
  et~al.
\newblock Spontaneous behaviour is structured by reinforcement without explicit
  reward.
\newblock \emph{Nature}, 614\penalty0 (7946):\penalty0 108--117, 2023.

\bibitem[Whittington et~al.(2020)Whittington, Muller, Mark, Chen, Barry,
  Burgess, and Behrens]{whittington2020tolman}
James~CR Whittington, Timothy~H Muller, Shirley Mark, Guifen Chen, Caswell
  Barry, Neil Burgess, and Timothy~EJ Behrens.
\newblock The tolman-eichenbaum machine: unifying space and relational memory
  through generalization in the hippocampal formation.
\newblock \emph{Cell}, 183\penalty0 (5):\penalty0 1249--1263, 2020.

\bibitem[Whittington et~al.(2025)Whittington, Dorrell, Behrens, Ganguli, and
  El-Gaby]{whittington2025tale}
James~CR Whittington, William Dorrell, Timothy~EJ Behrens, Surya Ganguli, and
  Mohamady El-Gaby.
\newblock A tale of two algorithms: Structured slots explain prefrontal
  sequence memory and are unified with hippocampal cognitive maps.
\newblock \emph{Neuron}, 113\penalty0 (2):\penalty0 321--333, 2025.

\bibitem[Wu(1983)]{wu1983convergence}
CF~Jeff Wu.
\newblock On the convergence properties of the em algorithm.
\newblock \emph{The Annals of statistics}, pages 95--103, 1983.

\bibitem[Nathan et~al.(1996)Nathan, Senior, and
  Subrahmonia]{nathan1996initialization}
Krishna Nathan, Andrew Senior, and Jayashree Subrahmonia.
\newblock Initialization of hidden markov models for unconstrained on-line
  handwriting recognition.
\newblock In \emph{1996 IEEE International Conference on Acoustics, Speech, and
  Signal Processing Conference Proceedings}, volume~6, pages 3502--3505. IEEE,
  1996.

\bibitem[Miller and Fumarola(2012)]{miller2012mathematical}
Kenneth~D Miller and Francesco Fumarola.
\newblock Mathematical equivalence of two common forms of firing rate models of
  neural networks.
\newblock \emph{Neural computation}, 24\penalty0 (1):\penalty0 25--31, 2012.

\end{thebibliography}

% References follow the acknowledgments in the camera-ready paper. Use unnumbered first-level heading for
% the references. Any choice of citation style is acceptable as long as you are
% consistent. It is permissible to reduce the font size to \verb+small+ (9 point)
% when listing the references.
% Note that the Reference section does not count towards the page limit.
% \medskip

% {
% \small

% [1] Alexander, J.A.\ \& Mozer, M.C.\ (1995) Template-based algorithms for
% connectionist rule extraction. In G.\ Tesauro, D.S.\ Touretzky and T.K.\ Leen
% (eds.), {\it Advances in Neural Information Processing Systems 7},
% pp.\ 609--616. Cambridge, MA: MIT Press.

% [2] Bower, J.M.\ \& Beeman, D.\ (1995) {\it The Book of GENESIS: Exploring
%   Realistic Neural Models with the GEneral NEural SImulation System.}  New York:
% TELOS/Springer--Verlag.

% [3] Hasselmo, M.E., Schnell, E.\ \& Barkai, E.\ (1995) Dynamics of learning and
% recall at excitatory recurrent synapses and cholinergic modulation in rat
% hippocampal region CA3. {\it Journal of Neuroscience} {\bf 15}(7):5249-5262.
% }

\newpage
\section*{NeurIPS Paper Checklist}

\begin{enumerate}

\item {\bf Claims}
    \item[] Question: Do the main claims made in the abstract and introduction accurately reflect the paper's contributions and scope?
    \item[] Answer: \answerYes{} % Replace by \answerYes{}, \answerNo{}, or \answerNA{}.
    \item[] Justification: We believe that the claims are well supported by the conceptual developments and numerical evaluations we made.

\item {\bf Limitations}
    \item[] Question: Does the paper discuss the limitations of the work performed by the authors?
    \item[] Answer: \answerYes{} % Replace by \answerYes{}, \answerNo{}, or \answerNA{}.
    \item[] Justification: A discussion of limitation is supplied at the end of the paper.

\item {\bf Theory assumptions and proofs}
    \item[] Question: For each theoretical result, does the paper provide the full set of assumptions and a complete (and correct) proof?
    \item[] Answer: \answerNA{} % Replace by \answerYes{}, \answerNo{}, or \answerNA{}.
    \item[] Justification: Our results do not contain proof-based theoretical results.

    \item {\bf Experimental result reproducibility}
    \item[] Question: Does the paper fully disclose all the information needed to reproduce the main experimental results of the paper to the extent that it affects the main claims and/or conclusions of the paper (regardless of whether the code and data are provided or not)?
    \item[] Answer: \answerYes{} % Replace by \answerYes{}, \answerNo{}, or \answerNA{}.
    \item[] Justification: Algorithms are stated in the appendix and we also make the code available.

\item {\bf Open access to data and code}
    \item[] Question: Does the paper provide open access to the data and code, with sufficient instructions to faithfully reproduce the main experimental results, as described in supplemental material?
    \item[] Answer: \answerYes{} % Replace by \answerYes{}, \answerNo{}, or \answerNA{}.
    \item[] Justification: Code is provided in a repository.

\item {\bf Experimental setting/details}
    \item[] Question: Does the paper specify all the training and test details (e.g., data splits, hyperparameters, how they were chosen, type of optimizer, etc.) necessary to understand the results?
    \item[] Answer: \answerYes{} % Replace by \answerYes{}, \answerNo{}, or \answerNA{}.
    \item[] Justification: We specify important parameters and settings in the text as well as provide the code for reproducibility.

\item {\bf Experiment statistical significance}
    \item[] Question: Does the paper report error bars suitably and correctly defined or other appropriate information about the statistical significance of the experiments?
    \item[] Answer: \answerNA{} % Replace by \answerYes{}, \answerNo{}, or \answerNA{}.
    \item[] Justification: We do not include experimental results that required significance testing.

\item {\bf Experiments compute resources}
    \item[] Question: For each experiment, does the paper provide sufficient information on the computer resources (type of compute workers, memory, time of execution) needed to reproduce the experiments?
    \item[] Answer: \answerYes{} % Replace by \answerYes{}, \answerNo{}, or \answerNA{}.
    \item[] Justification: In general, our experiments have very little compute requirement. The configurations used are stated in the appendix.
    
\item {\bf Code of ethics}
    \item[] Question: Does the research conducted in the paper conform, in every respect, with the NeurIPS Code of Ethics \url{https://neurips.cc/public/EthicsGuidelines}?
    \item[] Answer: \answerYes{} % Replace by \answerYes{}, \answerNo{}, or \answerNA{}.
    \item[] Justification: We have reviewed the code of ethics and can confirm that our work conforms with it.

\item {\bf Broader impacts}
    \item[] Question: Does the paper discuss both potential positive societal impacts and negative societal impacts of the work performed?
    \item[] Answer: \answerNA{} % Replace by \answerYes{}, \answerNo{}, or \answerNA{}.
    \item[] Justification: Our research aims to better understand task compositionality and how the brain may handle it. These do not have immediate relations to ML applications in the real world.
    
\item {\bf Safeguards}
    \item[] Question: Does the paper describe safeguards that have been put in place for responsible release of data or models that have a high risk for misuse (e.g., pretrained language models, image generators, or scraped datasets)?
    \item[] Answer: \answerNA{} % Replace by \answerYes{}, \answerNo{}, or \answerNA{}.
    \item[] Justification: See previous question.

\item {\bf Licenses for existing assets}
    \item[] Question: Are the creators or original owners of assets (e.g., code, data, models), used in the paper, properly credited and are the license and terms of use explicitly mentioned and properly respected?
    \item[] Answer: \answerYes{} % Replace by \answerYes{}, \answerNo{}, or \answerNA{}.
    \item[] Justification: We provide citations whenever necessary.

\item {\bf New assets}
    \item[] Question: Are new assets introduced in the paper well documented and is the documentation provided alongside the assets?
    \item[] Answer: \answerYes{} % Replace by \answerYes{}, \answerNo{}, or \answerNA{}.
    \item[] Justification: We provide code used for our experiments but at the current moment do not expect them to be used as packages.

\item {\bf Crowdsourcing and research with human subjects}
    \item[] Question: For crowdsourcing experiments and research with human subjects, does the paper include the full text of instructions given to participants and screenshots, if applicable, as well as details about compensation (if any)? 
    \item[] Answer: \answerNA{} % Replace by \answerYes{}, \answerNo{}, or \answerNA{}.
    \item[] Justification: No crowdsourcing nor research with human subjects was involved.

\item {\bf Institutional review board (IRB) approvals or equivalent for research with human subjects}
    \item[] Question: Does the paper describe potential risks incurred by study participants, whether such risks were disclosed to the subjects, and whether Institutional Review Board (IRB) approvals (or an equivalent approval/review based on the requirements of your country or institution) were obtained?
    \item[] Answer: \answerNA{} % Replace by \answerYes{}, \answerNo{}, or \answerNA{}.
    \item[] Justification: See previous question.

\item {\bf Declaration of LLM usage}
    \item[] Question: Does the paper describe the usage of LLMs if it is an important, original, or non-standard component of the core methods in this research? Note that if the LLM is used only for writing, editing, or formatting purposes and does not impact the core methodology, scientific rigorousness, or originality of the research, declaration is not required.
    %this research? 
    \item[] Answer: \answerNA{} % Replace by \answerYes{}, \answerNo{}, or \answerNA{}.
    \item[] Justification: LLMs were only used for writing and editing.

\end{enumerate}

%%%%%%%%%%%%%%%%%%%%%%%%%%%%%%%%%%%%%%%%%%%%%%%%%%%%%%%%%%%%

\appendix
% change figure counter to be appendix fig and reset to 0
\setcounter{figure}{0}
\renewcommand\thefigure{S.\arabic{figure}}   
{\LARGE Separating the \textit{what} and \textit{how} of compositional
computation to enable reuse and continual learning \vspace{0.2cm}\\ Appendix}

All our codes are available on the  \href{https://github.com/dunckerlab/contextrnn}{public repository}.
\section{Formalization of compositional tasks and task designs}
\label{appendix: task design}
We propose a formalization of shared compositional structures in sequence-to-sequence (seq2seq) tasks and show that the formulation is expressive enough to capture many commonly used tasks in the literature. In each trial of a seq2seq task, the learner receives an input time sequence $\vec{s}_{t=1,...,T}$ and needs to produce a target output sequence $\vec{y}_{t=1,...,T}$. Different tasks (indexed by $c\in \mathbb{Z}^+$) are distinguished from each other by their input distributions as well as the underlying input-to-target rules, together encoded in $p(\vec{s}_{1:T}, \vec{y}_{1:T}|c)$. For notational convenience let $\vec{q}_t\equiv [\vec{s}_t,\vec{y}_t] \in \mathbb{R}^{D_q}$.

We model this distribution over task-observables by introducing a set of latent variables resulting in the joint distribution
\begin{equation}
   p(\vec{q}_{1:T}, z_{1:T}, x, c)= p(x|c) p(z_1|c) \prod_{t=1}^{T-1}  p(z_{t+1}| z_{t}, c)\prod_{t=1}^T p(\vec{q}_t|z_t,x) \; p(c).
   \label{eq:model equation}
\end{equation}
We initially provide intuition about our modeling choices, before formally specifying the distributions for the tasks of the main paper in section \ref{appendix subsec: task generation model} and \ref{appendix subsec: task generation details} below.

To model different tasks with a shared compositional structure, we first observe that in many tasks used in neuroscience experiments, each trial can be temporally segmented into discrete epochs, which we denote as $z_t\in\mathbb{Z}^+$. 
Each epoch has its distinctive input and target statistics, though they may exhibit complex temporal dependencies across epochs throughout the trial. 
To illustrate this, we can consider a simple task testing the learner's ability to memorize: During a stimulus epoch, inputs provide information about a trial-specific latent variable (e.g., a dot on the screen indicating an angle $\theta$). The inputs eventually turn off, and the learner needs to maintain the relevant information in memory until a cue solicits a $\theta$-dependent response (e.g., a saccade in the direction of the angle). This example corresponds to the \texttt{MemoryPro} task from the main paper.

One might initially attempt to model this epoch dependence as a simple Hidden Markov Model (HMM), where each epoch has its own observation model $p(\vec{q}_t|z_t)$. However, this approach would fail to capture that the $\vec y_t$ during the response epoch is coupled to the $\vec{s}_t$ from the stimulus epoch via the shared trial-specific latent variable. In our example above, the statistics of the stimulus epoch and response epoch are coupled since they are both dependent on $\theta$. To take this cross-epoch dependency into account, we explicitly model the ``stimulus condition'', indexed by the trial variable $x$. In the specific example above, there may be a list of possible directions and $x$ selects one to use as $\theta$. Thus, in our model of cognitive tasks, the observed inputs and target responses depend not only on the latent epoch $z_t$ but also a second latent variable $x$, which additionally parameterizes the observation models of all epochs. 
% The introduction of $x$ makes our model (for a single task $c$) a special case of an HMM whose emissions are another mixture model.

We model tasks as being compositionally related if we can use the same underlying set of task epochs and conditional distributions over $\vec q_t | z_t, x$ to describe them.
For a simple example of different tasks sharing compositional structure, we can consider the  \texttt{MemoryPro} task described above, and construct a second task, \texttt{MemoryAnti}, which shares the same stimulus epoch but requires a response in the opposite direction of $\theta$. Here, the dependence of the response epoch on $\theta$ is different (and \texttt{MemoryAnti} will therefore have a different response epoch relative to  \texttt{MemoryPro}), but the statistics of the stimulus epoch are identical across the two tasks.
The differences in what epochs contribute to each task and how one epoch transitions to the next is captured via task-dependent Markovian transitions over epochs $p(z_{t+1}|z_t,c)$.

\subsection{Generative model}
\label{appendix subsec: task generation model}

To make our model of neuroscience tasks tractable and applicable to training RNNs, we make some simplifying choices for its components. For all tasks, $p(x|c)$ is assumed to be a uniform distribution over some finite set of size $N_x$. $p(z_{1:T}|c)$ is assumed to be a Markov process with the initial state distribution and transition probabilities determined by $c$. Finally, the observation model $p(\vec{q}_t|z_t,x)$ is assumed to be a multivariate Gaussian with a $(z_t,x)$-dependent mean. Under these assumptions, our model can be seen as an HMM with $c$-dependent latent-state dynamics and $x$-dependent emission models for all latent states. If there is only one task and one possible $x$ value in all trials, the model reduces to a standard HMM. In the more general case, it can be viewed as a mixture (across tasks identities) of HMMs with Gaussian Mixture emissions (across trial variables $x$) .

In summary, for trials $r=1,\dots, N_{trials}$ we generate tasks as
\begin{align*}
c_r & \sim p(c) && \text{task identity} \qquad c_r \in \{ 1, \dots, N_c\} \\
z_1^r|c_r & \sim \text{Discrete}(\vec \Pi^c) && \text{epoch identity} \qquad z_t^r \in \{ 1, \dots, N_z\} \\
z_t^r | z^r_{t-1}=z', c_r=c & \sim \text{Discrete}(\vec \Lambda^c_{:,z'}) \\
x_r|c_r & \sim \text{Discrete}(\frac{1}{N_x} \vec 1) && \text{trial variable} \qquad x_r \in \{ 1, \dots, N_x\} \\
\vec q_t^r | z_t^r=z, x_r=x & \sim \mathcal{N}\left( \bar{\vec q}_{z,x}, \sigma^2 I \right) && \text{inputs and targets} \qquad \vec q_t^r \in \mathbb{R}^d
\end{align*}

This model thus allows us to specify a set of $N_{c}$ tasks sharing a compositional structure as follows. Let $N_z$ denote the number of epochs shared among these tasks. The composition of the $c$-th task is specified by its epoch-transition parameters: the transition probabilities $\vec {\Lambda}^c \in \mathbb{R}^{{N_z} \times N_z}:\Lambda^c_{z,z'}=p(z_t=z|z_{t-1}=z',c)$ and initial probabilities $\vec\Pi^c \in \mathbb{R}^{N_z}:\Pi^c_z=p(z_1=z|c)$. The observation model of each epoch is specified by the means corresponding to different $x$ values, $\{\bar{\vec{q}}_{z,x} \in \mathbb R^{D_q}\}_{x=1,...,N_x} $, where $\bar{\vec{q}}_{z,x}$ is the mean of the multivariate Gaussian $p(\vec{q}_t|z_t=z,x)$. Thus, altogether an ensemble of $N_c$ tasks composed from $N_z$ epochs is specified by the tuple $(\{\bar{\vec{q}}_{z,x}\}_{z=1,...,N_z;x=1,...,N_x},\{(\vec{\Lambda}^c,\vec{\Pi}^c)\}_{c=1,...,N_c})$, where the first component specifies the epochs and the second specifies how they are used to compose the tasks. 

\subsection{Expressing common cognitive and motor tasks in our framework}
\label{appendix subsec: task generation details}

Below, we provide explicit details of the distributions and parameters used to generate our task set. We follow the same epoch notation introduced in Table \ref{tab:tasks} of the main paper. We initially describe the temporal structure in terms of epochs of each task, and then define epoch-specific distributions.

All tasks start with the \textit{fixation} epoch ($F$), which presents no stimulus and an active fixation cue. The required output is to maintain fixation without response. In all the non-response epochs (\textit{stimulus} $S$, \textit{memory} $M$ and \textit{decision stimuli} $S_{DM}$), the fixation cue is on and the required output is to maintain fixation without producing a response. During the response epochs ($R_P,R_A,R_{M,P},R_{M,A},R_{DM,P},R_{DM,A}$), the fixation cue turns off and the learner needs to stop fixation and produce a response $\phi$ according to some rule, as described below. Without loss of generality, we considered 8 possible stimulus conditions per task ($N_x=8$) and generate $x$ i.i.d. from a uniform distribution for the 8 possible values.

\textbf{Delayed response tasks (\texttt{DelayPro}, \texttt{DelayAnti})}. %
After $F$, the stimulus epoch ($S$) presents an angle $\theta$, chosen from $\{0,\pi/4,...,7\pi/4\}$ depending on $x$. The stimulus presentation stays on during the ensuing response epoch ($R_P$ for \texttt{DelayPro} or $R_A$ for \texttt{DelayAnti}), where the target output becomes $\phi=\theta$ ($R_P$) or $\phi=\theta+\pi$ ($R_A$). 
      
\textbf{Memory-guided response tasks (\texttt{MemoryPro}, \texttt{MemoryAnti})}. %
After $S$, stimulus presentation disappears in the memory epoch ($M$). During the ensuing response epoch ($R_{M,P}$ for \texttt{MemoryPro}, $R_{M,A}$ for \texttt{MemoryAnti}), there is still no stimulus presentation and the learner must produce a response based on the memorized $\theta$: $\phi=\theta$ ($R_{M,P}$) or $\phi=\theta+\pi$ ($R_{M,A}$).
      
\textbf{Decision making tasks (\texttt{DMPro}, \texttt{DMAnti})}. %
After $F$, a decision stimuli epoch ($S_{DM}$) presents two stimuli simultaneously. $\theta=0$ with strength $\gamma$ in input dims 1, 2 and $\theta=\pi$ with strength $\gamma'$ in input dims 3, 4. The strengths $(\gamma,\gamma')$ are determined from a set of pairs by $x$ and scale the input channels. During the ensuing response epoch ($R_{DM,P}$ for \texttt{DMPro}, $R_{DM,A}$ for \texttt{DMAnti}), the stimuli persist and the required output $\phi$ is the direction of the stimulus with a higher strength ($R_{DM,P}$) or the one with a lower strength ($R_{DM,A}$).

The epoch structure of the different tasks is encoded via $\vec \Lambda^c$. Altogether, the 6 tasks can be described as compositions using a shared pool of 10 epochs. Given the epoch identity, the conditional distributions of the inputs and responses are generated as follows:

\begin{center}
    \begin{tabular}{l|l|l}    
      epoch &  $\bar{\vec s}_{z,x}$ & $\bar{\vec y}_{z,x}$ \\
       \hline
       $F$, $M$ & $[0, 0, 0, 0, 0]$ & $[0, 0, 0]$\\ 
       $S$ & $[\cos \theta, \sin \theta, 0, 0, 0]$ & $[0, 0, 0]$\\
%       $M$ & $[0, 0, 0, 0, 0]$ & $[0, 0, 0]$\\
       $R_P$ & $[\cos \theta, \sin \theta, 0, 0, 1]$ & $[\cos \theta, \sin \theta, 1]$\\
       $R_A$ & $[\cos \theta, \sin \theta, 0, 0, 1]$ & $[\cos (\theta + \pi), \sin (\theta + \pi), 1]$\\
       $R_{M,P}$ & $[0, 0, 0, 0, 1]$ & $[\cos \theta, \sin \theta, 1]$\\
       $R_{M,A}$ & $[0, 0, 0, 0, 1]$ & $[\cos (\theta + \pi), \sin (\theta + \pi), 1]$\\
       $S_{DM}$ & $[\gamma \cos \theta,  \gamma \sin \theta,  \gamma' \cos \theta',  \gamma' \sin \theta', 1]$ & $[1, 0, 0]$\\
       $R_{DM,P}$ & $[\gamma \cos \theta,  \gamma \sin \theta,  \gamma' \cos \theta',  \gamma' \sin \theta', 1]$ & $[\cos \phi, \sin \phi, 1]$, where $\phi = \vec{1}_{\gamma > \gamma'}\theta + \vec{1}_{\gamma' > \gamma}\theta'$\\
       $R_{DM,A}$ & $[\gamma \cos \theta,  \gamma \sin \theta,  \gamma' \cos \theta',  \gamma' \sin \theta', 1]$ & $[\cos \phi, \sin \phi, 1]$, where $\phi = \vec{1}_{\gamma < \gamma'}\theta + \vec{1}_{\gamma' < \gamma}\theta'$\\ \hline
    \end{tabular}
\end{center}
Where $\vec{1}$ is an indicator that takes on value 1 if the subscript is true and 0 otherwise. This fully specifies the conditional distributions $p(\vec{q}_t|z_t,x)$ for each task epoch with $\bar{\vec q}_{z,x} = [\bar{\vec s}_{z,x}, \; \bar{\vec y}_{z,x}]$. For the \texttt{DM} tasks, we restrict the stimulus values to two locations $\theta=0, \theta'=\pi$, but pick $N_x=8$ different combinations of possible $(\gamma,\gamma')$ pairs. Note that while we generate the input and response distributions using stimulus values $\theta$ to follow convention from the literature \citep{yang2019task, driscoll2024flexible}, each value of $\theta$ (or $(\gamma,\gamma')$ in the \texttt{DM} tasks) maps onto a different $x$ value. For $\sigma$ we used 0.05; for ($\gamma,\gamma'$), $x$ selects from $[0.5, 1], [1, 2], [0.5, 2], [0.2, 1.5],[1, 0.5], [2, 1], [2, 0.5], [1.5, 0.2]$.

Note that while we have followed many conventions from previous work in the task design, these previous approaches tend to implement each task individually but with related distributional assumptions \cite{yang2019task,driscoll2024flexible,duncker2020organizing,costacurta2024structured}. Modeling shared structure across tasks through an explicit, shared generative framework for an entire task family is novel. While this was not the focus of the main paper, it is worth noting that access to a description of shared statistical structure in the input and target response pairs of each task forms an important baseline for expectations of shared statistical structure across tasks in the solution emerging after RNN training \cite{yang2019task, driscoll2024flexible, duncker2020organizing, riveland2024natural}.

\section{Online learning and inference of compositional task structures}
\label{appendix: learning algo}

\subsection{Online learning and inference}
\label{appendix subsec: contextual learning and inference}
In this section, we provide additional details on the algorithms developed for performing posterior inference over the latent variables of the task model, and online (one-trial-at-a-time) learning of the task model.
\paragraph{Inference.} We can perform exact inference by utilizing a message passing scheme similar to that used for performing inference in classic HMM models. 

Let $\alpha_t^r(z, x, c) = p(\vec q^r_{1:t}, z_t^r=z|x_r=x, c_r=c)$ denote the forward (filtering) message for a given task. During a filtering pass, we compute
\begin{align}
    \alpha_1^r(z, x, c) & = p(z_1^r=z| c_r=c)p(\vec q^r_1|z_1^r=z, x_r=x) \\
    \alpha_{t+1}^r(z, x, c) &= \left(\sum_{z'=1}^{N_z}\alpha_t^r(z', x, c)p(z_{t+1}^r=z|z_t^r=z', c_r=c)\right)
p(\vec q^r_{t+1}|z_{t+1}^r=z, x_r=x)
\end{align}
Marginalizing the forward message at the final time-step allows us to compute the marginal likelihood over observations
\begin{align}
    p(\vec q^r_{1:T}|x_r=x, c_r=c) & = \sum_{z=1}^{N_z} \alpha_T^r(z, x, c) \\
\end{align}
Let $\beta^r_t(z, x, c) = p(\vec q^r_{(t+1):T}|z_t^r=z, x_r=x, c_r=c)$ denote the backwards (smoothing) message. During the smoothing pass, we compute
\begin{align}
    \beta^r_T(z, x, c) & = 1 \\
    \beta^r_t(z, x, c) & = \sum_{z'=1}^{N_z}p(z^r_{t+1}=z'|z^r_t=z, c_r=c)p(\vec q^r_{t+1}|z^r_{t+1}=z', x_r=x)\beta^r_{t+1}(z', x, c)
\end{align}
Given these quantifies, we can compute the joint posterior over $z^r_t$, $x_r$ and $c_r$ as
\begin{align}
    &\gamma^r_t(z, x, c) = p(z^r_t=z, x_r=x, c_r=c|\vec q^r_{1:T}) \\
& =\frac{p(\vec q^r_{1:t}, z^r_t=z|x_r=x, c_r=c)p(\vec q^r_{(t+1):T}|z^r_t=z, x_r=x, c_r=c)p(x_r=x, c_r=c)}{\sum_{x, c} p(\vec q^r_{1:T}|x_r=x, c_r=c)p(x_r=x, c_r=c)} \\
& = \frac{\alpha^r_t(z, x, c)\beta^r_t(z, x, c)p(x_r=x, c_r=c)}{\sum_{x, c}\sum_{z'=1}^{N_z}\alpha^r_T(z', x, c)p(x_r=x, c_r=c)}
\end{align}
Note that the filtering and smoothing passes can be done for each value of $x$ and $c$ in parallel. 
Finally, for later use in the online learning algorithm we also compute the joint posterior over $z^r_{t-1}$, $z^r_t$, $x_r$ and $c_r$ as 
\begin{align}
&\xi^r_t(z, z', x, c) = p(z^r_{t-1}=z,z^r_{t}=z', x_r=x, c_r=c|\vec q^r_{1:T})\\
&=\frac{\alpha^r_{t-1}(z, x, c) p(z^r_t=z'|z^r_{t-1}=z,c_r=c) p(\vec q^r_{t}|z^r_t=z',x_r=x) \beta^r_{t}(z', x, c) p(x_r=x, c_r=c)}{\sum_{x, c} p(\vec q^r_{1:T}|x_r=x, c_r=c)p(x_r=x, c_r=c)}
\end{align}

\paragraph{Online learning.}
Learning aims to recover parameters of the generative model, $(\{\bar{\vec{q}}_{z,x}\}_{z=1,...,N_z;x=1,...,N_x},\{(\vec{\Lambda}^c,\vec{\Pi}^c)\}_{c=1,...,N_c})$, up to a permutation over $z$ and $x$. %For the particular set of tasks we considered, since $F,M$ epochs have identical $\bar {\vec q}_{z,x}$ for all $x$, they are indistinguishable and will be combined by the learning algorithm. Thus, instead of learning the 10 epochs in the generative process for our set of tasks, it will end up learning 9 epochs, including a combined $F/M$ epoch. In terms of the transition matrices, this creates a complication for \texttt{MemoryPro, MemoryAnti} tasks where both $F$ and $M$ epochs appear in each trial. The learned transitions will not be deterministic in the sense that the $F/M$ epoch may transition to either the stimulus epoch or the response epoch. The `ground-truth' parameters used for plotting in Fig. \ref{fig:task-model-learning} and Fig. \ref{appendix fig: task learning with different orders} refer to the optimal parameters with a merged $F/M$ epoch.
Performing parameter learning with EM requires computing the expected counts of visiting particular epochs or transitioning across epochs across all trials. When all trials are available as a batch, this involves sums over the quantities computed during inference for each trial. To make notation more compact, we denote $X$ as the set of sufficient statistics needed to update the set of model parameters $\Theta$. Let $\Theta^{(i,k)}$ denote the learned parameters after seeing the $i$-th trial and running $k$ EM iterations. Let $X^{r,(i,k)}$ denote the single-trial statistics on trial $r$, computed using parameters $\Theta^{(i,k)}$.
When the learning algorithm has access to all trials throughout learning (batch EM), the updates take the form
\begin{align}
    \Theta^{(i,k+1)} & = f(S^{(i,k)}(X)) \quad S^{(i,k)}(X) = \sum_{r=1}^i X^{r,(i,k)}.
\end{align}
For example, for the parameter estimates for the transition matrix, this takes the form
\begin{align}
% \pi_{jm}^{i, k+1} & =\frac{\sum_{l}\Gamma^{i, k}_{jlm}(0)}{\sum_{j'm'}\sum_{l}\Gamma^{i,k}_{j'lm'}(0)} = \frac{\Gamma^{i, k}_{j\cdot m}(0)}{\Gamma^{i, k}_{\cdot \cdot \cdot}(0)} \\
\Lambda_{czz'}^{i,k+1} & =\frac{\sum_{t=1}^{T} \sum_{x=1}^{N_x} \Xi^{i,k}_{cx zz'}(t)}{\sum_{t=1}^{T} \sum_{x=1}^{N_x} \sum_{z''=1}^{N_z} \Xi^{i,k}_{cxz''z'}(t)}.%\\
% \bar{\vec q}_{ml}^{i,k+1} & =\frac{\sum_{t=0}^{T}\Gamma^{i, k}_{\cdot lk}(t)\vec q_{t}}{\sum_{t=0}^{T}\Gamma^{i, k}_{\cdot lk}(t)}
\end{align}
with
% $\Theta^{k}$ denote the set of parameter estimates after the $k$th iteration of learning and introduce a  superscript $r$ to index variables on different trials. Define the sum of expected sufficient statistics up to trial $i$ on iteration $k$ as
% %
\begin{align}
%  \gamma^c_t(i, k) &= p(z_t=i|\vec q_{1:T}, x=k, c) & \equiv 
% \gamma^{i,k}\left(c^i,x^i,z^i_{t}\right)&\equiv p\left(c^i,x^i,z^i_{t}|w^i_{0\rightarrow T};\Theta^{k}\right) = \gamma^{i,k}_{cxz}(t)\\
% \xi^{i,k}\left(c^i,x^i,z^i_{t-1},z^i_{t}\right)& \equiv p\left(c^i,x^i,z^i_{t-1},z^i_{t}|w^i_{0\rightarrow T};\Theta^{k}\right) = \xi^{i,k}_{cxzz'}(t)\\
% \Gamma^{i,k}_{cxz}(t) & = \sum_{r=1}^i p\left(c_r=c,x_r=x,z^r_{t}=z|\vec q^r_{1:T};\Theta^{k}\right)\\
\Xi^{i,k}_{cxzz'}(t) & = \sum_{r=1}^i p\left(c_r=c,x_r=x,z^r_{t-1}=z,z^r_{t}=z'|\vec q^r_{1:T};\Theta^{k}\right).
\end{align}
% Given these quantities, the exact (batch) parameter updates will be of the form
% 
In general, the batch EM parameter updates take well-known forms for HMMs and GMM-HMMs, which is why we only give an example here in the interest of brevity.

When trials are only available one-trial-at-a-time and cannot be revisited (online EM), the sums of expected sufficient statistics across trials have to be approximated and updated after each trial instead, leading to an approximation to the batch updates. We propose the following update rule to perform parameter learning online
\begin{align}
    &\Theta^{(i,k+1)} = (1-\eta_{params}) \Theta^{(i-1,K)} + \eta_{params} f(S^{(i,k)}_{online}(X)). \\
    & S^{(i,k+1)}_{online}(X) = (1-\eta_{stats}) S^{(i-1,K)}_{online}(X) + X^{i,(i,k)}.
\end{align}
where $K$ denotes the number of iterations per trial. After learning the $i$-th trial, we only need to store $\Theta^{(i,K)}$ and $S^{(i,K)}_{online}(X)$, taking the form
\begin{align}
    &\Theta^{(i,K)} = \sum_{r}^i (1-\eta_{params})^{i-r} \eta_{params} f(S^{(r,K)}_{online}(X)) + (1-\eta_{params})^{i+1} \Theta_{init.} \\
    &S_{online}^{(i,K)}(X) = \sum_{r}^i (1-\eta_{stats})^{i-r} X^{r,(r,K)}.
\end{align}

\begin{figure}
    \centering
    \includegraphics[width=0.8\linewidth]{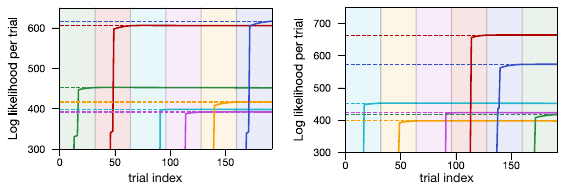}
    \caption{Online continual learning of task structures with some example task orderings. See Fig.\ref{fig:task-model-learning} for legends.}
    \label{appendix fig: task learning with different orders}
\end{figure}

We summarize this online learning approach in Algorithm \ref{alg:cap} and show performance for different training orders (supplementing Figure \ref{fig:task-model-learning} in the main paper) in Figure \ref{appendix fig: task learning with different orders}.

\begin{algorithm}
\caption{High-level Overview of the Incremental EM algorithm. }\label{alg:cap}
\begin{algorithmic}
\State $\Theta \gets \text{init.}$ \Comment{Initialize parameter estimates.}
\State $S(X) \gets \mathbf{0}$ \Comment{These are the estimated sums of sufficient statistics across trials}
\For{$r=1,...,N_{trials}$}
    \State $\Theta \gets incremental\_initialize(\Theta,q^r_{1:T_r},c_r)$ \Comment{See \ref{appendix subsec: online learning initialization}.}
    \State $\hat{\Theta} \gets \Theta$  \Comment{Create a temporary copy of the parameters}
    \State $\hat{S}(X) \gets \text{None}$ 
    \For{$k=1,...,K$}
        \State $X \gets {getstats}(\vec q^r_{1:T_r},c_r,\hat{\Theta}) $ \Comment{Compute sufficient statistics of this trial with current params.}
        \State $\hat{S}(X) \gets (1-\eta_{stats})S(X) + X$ \Comment{Decay the sums from the previous trial and add new stats}
        \State $\hat{\Theta} \gets \Theta \odot (1-\eta_{params}M_{gate}) + \eta_{params} f(\hat{S}(X)) \odot M_{gate}$ \Comment{Incremental update of params. using the modified sums. $M_{gate}$ is a binary mask controlling which parameters are updated (see \ref{appendix subsec: gated updates}).}
    \EndFor
    \State $\Theta \gets \hat \Theta$
    \State $S(X) \gets \hat{S}(X)$
\EndFor
\end{algorithmic}
\end{algorithm}

\subsection{Incremental initialization of model parameters}
\label{appendix subsec: online learning initialization}

Even for simple models such as the Gaussian Mixture Model, EM is susceptible to local optima \citep{wu1983convergence}. To avoid convergence to bad local optima, it is important to obtain good initializations for the model parameters.
For simple time-series models such as the HMM, this is typically achieved by collapsing the sequences across time, performing clustering (e.g. with K-means \citep{nathan1996initialization}), and using the resulting cluster centers as the initial observation-model parameters, given each latent discrete state. 

In our model, we need to initialize the underlying cluster centers of the observation model, $\{ \hat {\vec{q}}_{z,x} \}_{z=1,...,\hat{N_z},x=1,...,N_x}$. Here, $\hat{N_z}$ reflects the fact that the total number of epochs in the entire dataset is unknown \emph{a priori}. We overcome this by setting up a large number of `slots' ($\hat{N_z}\geq N_z$), exceeding the likely total number of epochs in the task family. 
Our setting adds two significant challenges relative to simple HMMs:
%
%Our setting is significantly challenging for two reasons. 
First, each trial contains only a few epochs and a specific stimulus condition ($x$ value). Therefore, clustering each trial can only initialize a small subset of the $N_x  N_z$ mixture components of the observation model. 
Second, since there is one stimulus condition $x$ per trial, the mixture components $\bar {\vec{q}}_{z,x}$ explaining a given trial must be allocated to the same $x$, but different epoch states $z$. Initialization schemes based on simple clustering approaches on the entire trial (such as that outlined for HMMs above) would be agnostic to this structure and fail to provide a feasible initial set of parameters.
To overcome these issues, we introduce an `incremental initialization' scheme, which is applied to all the estimated means $\{ \hat {\vec{q}}_{z,x} \}$ before learning each trial.

To introduce the scheme, we first introduce the notion of the putative $z,x$, denoted $\tilde{z},\tilde{x}$. This is to highlight that learning the correct $\{\bar{q}_{z,x}\}$ does not require inferring the real $z,x$ in the generative process but only requires them to be correct up to a permutation. Let $\tilde{N_z},\tilde{N_c}$ denote the number of epochs and task slots in the learning algorithm, respectively. Note that these do not need to be set as the correct $N_z,N_c$ -- they can simply be large integers. We do assume that the system knows the correct number of $x$ values, $N_x$. The learner keeps track of $\{ \hat {\vec{q}}_{\tilde z, \tilde x} \}_{\tilde{z}=1,...,\tilde{N_z},\tilde{x}=1,...,\tilde{N_x}}$ as well as two Boolean-valued tables, $F_{c,x}$ of size $\tilde{N_c} \times N_x$ and $F_{x,z}$ of size $\tilde{N_x} \times \tilde{N_z}$. The two tables keep track of which combinations of $c,\tilde{x}$ and $\tilde{x},\tilde{z}$ have been encountered.

When the $r$th trial is observed and the learner has access to $(\vec{q}^r_{1:T_r},c_r)$, the scheme is threefold: 
\begin{enumerate}
\item We perform K-means clustering on the entire sequence $\vec{q}^r_{1:T_r}$. This gives us a set of cluster means. Since we assume the epochs to have piecewise constant inputs and mild noise, the means correspond to the different epochs that appeared in this trial. The challenge now is to assign $\tilde{z},\tilde{x}$ to these means, and to use them to initialize $\hat{\vec{q}}_{\tilde{z},\tilde{x}}$ accordingly. 
\item All cluster means from step (1) should be assigned the same $\tilde{x}_r$. We check the means against $\{ \hat {\vec{q}}_{\tilde z, \tilde x} \}$ and decide on $\tilde{x}_r$ according to a set of rules and $F_{c,x}$. The $c_r,\tilde{x}_r$ pair is marked `encountered' in $F_{c,x}$.
\item Given $\tilde{x}_r$ from step (2), we treat the cluster means not found in $\{ \hat {\vec{q}}_{\tilde z, \tilde x} \}_{\tilde{x} = \tilde{x}_r}$ as unfamiliar $\tilde{z},\tilde{x}$, meaning that they represent previously unseen $\tilde{z},\tilde{x}$ combinations and should be used to initialize. Each center is assigned a different $\tilde{z}$ that has not been encountered (according to $F_{x,z}$). The $\tilde{z},\tilde{x}_r$ pairs are marked as `encountered' in $F_{x,z}$.
\end{enumerate}
%
% First, it performs K-means clustering of $\vec{q}_{1:T}$. Second, it infers a $\tilde{x}$ for this trial by checking whether any of the $\tilde{x}$ previously encountered for this task can adequately `explain' the clusters. This is done by checking the cluster centers against $\{ \tilde {\vec{q}}_{\tilde z, \tilde x} \}$ for different $\tilde{x}$ and deciding according to a fixed set of rules. Finally, for the inferred $\tilde{x}$, it treats the cluster centers not found in $\{ \tilde {\vec{q}}_{\tilde z, \tilde x} \}$ as unfamiliar $\tilde{z},\tilde{x}$ combinations and initializes $\{ \tilde {\vec{q}}_{\tilde z, \tilde x} \}$ accordingly.
%

\subsection{Gated updates to parameters}
\label{appendix subsec: gated updates}
Since each trial contains information about only one task and the few epochs it is associated with, it does not make sense to update parameters related to other tasks and epochs. For $\vec \Lambda^c$, we simply gate it such that only the transition matrix corresponding to the current task (the label of which is given) is updated. For $\{ \hat {\vec{q}}_{z,x} \}$, we infer which epochs appeared in this trial using the posterior $p(z_{1:T}^r|\vec q^r_{1:T})$. Only epochs with a sufficiently high chance of appearance have their $\{ \hat {\vec{q}}_{z,x} \}$ updated. 

\subsection{Epoch identifiability in our set of tasks}
For the particular set of tasks we considered, since $F,M$ epochs have identical $\bar {\vec q}_{z,x}$ for all $x$, they are indistinguishable. Thus, our learning algorithm will combine them into a single epoch, which we denote as $F/M$. Instead of learning the 10 epochs in the generative process for our set of tasks, the task-model will end up learning 9 epochs, including the combined $F/M$ epoch. In terms of the transition matrices, this creates a complication for \texttt{MemoryPro, MemoryAnti} tasks where both $F$ and $M$ epochs appear in each trial. The learned transitions will not be deterministic in the sense that the $F/M$ epoch may transition to either the stimulus epoch or the response epoch. The `ground-truth' parameters used for plotting in Fig. \ref{fig:task-model-learning} and Fig. \ref{appendix fig: task learning with different orders} refer to the optimal parameters with a merged $F/M$ epoch.
In future work, it will be interesting to investigate how epochs that are computationally distinct (e.g. holding still in $F$ vs. holding still while maintaining a memory in $M$) but map onto the same observations may be distinguished, e.g. via feedback from the downstream network implementing the different computations for each epoch.

\section{RNN architecture and hyperparameter settings}
\subsection{Default RNN architecture and task parameters}
\label{appendix subsec: RNN architecture}
\begin{center}
\renewcommand{\arraystretch}{1.5}
\begin{tabular}{|c|c|}
\hline
\textbf{Parameter} & \textbf{Value}  \\
\hline
$\alpha$   &  0.1    \\
\hline
$\sigma_r$   & 0.05  \\
\hline
$\phi$  & ReLU \\
\hline
number of hidden units & 256 \\
\hline
rank of $U_z, V_z$ & 3 \\
\hline
input noise std & $\sqrt{2/\alpha}\ \sigma_{in}$, where $\sigma_{in}=0.01$\\
\hline
minimum duration of a epoch & 5 time steps \\
\hline
$p(z_{t+1} = z_{t} | c)$ & 0.9\\
\hline
\end{tabular}
\end{center}

\subsection{Default training protocol}
We used a batch size of 256 and trained each task for 1000 batches unless otherwise specified. 

For the context-modulated RNN, the learning rate for the weights associated with each context $z$ was initially set to $\eta_z = 0.001$. After training each task $c$, $\eta_z$ was multiplied by a decay factor $\gamma = 0.5$ for any context with $p(z \mid c) > 0.001$. During training on task $c$, $L_2$ weight regularization was applied with a coefficient of $10^{-5}$ for contexts with $p(z \mid c) > 0.001$, and set to 0 for all other contexts.

For baseline algorithms using general ("vanilla") RNN architectures, the learning rate was set to 0.01, and the $L_2$ weight regularization coefficient was set to $10^{-5}$. Parameter choices were determined by a coarse grid search.

\subsection{Loss function and performance measure}
\label{appendix subsec: loss and performance}
The loss function is a weighted mean square error similar to \cite{yang2019task, driscoll2024flexible}. $L:=\langle m_{i,t} (y_{i, t}-\hat{y}_{i, t})^2 \rangle_{i, t}$, where $i$ is the index of the output units, $m_{i, t} = 1$ for response epochs and $m_{i, t} = 0.2$ for all other epochs.

A trial is considered correct if the network maintained fixation for all time steps before the fixation cue turns off, and responded to the correct direction for time steps in the response epoch. If the activity of the fixation output exceeds 0.5, the network is considered to have broken fixation. The network's response direction is considered correct if its angular difference from the target direction is less than $\pi/10$. Average performance and test loss were calculated on 200 test trials for each task.

\section{Supplementary results}
\label{appendix: suppl results}

\subsection{Additional results on transfer learning}
\begin{figure}
\centering
\includegraphics{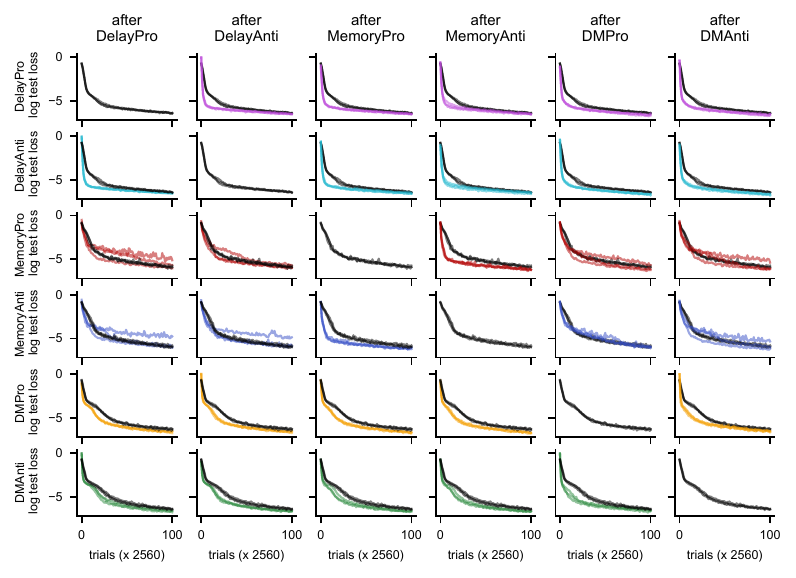}
% \vspace{-2ex}
\caption{Additional results on forward transfer. Colored curves show the log test loss of task B (indicated by the row label) when trained after task A (column label). For comparison, black curves in each row show the log test loss of task B when trained from scratch. We plot results across three random seeds for each task order.}
\label{fig:forward_transfer_supp}
\end{figure}

\begin{figure}
\centering
\includegraphics{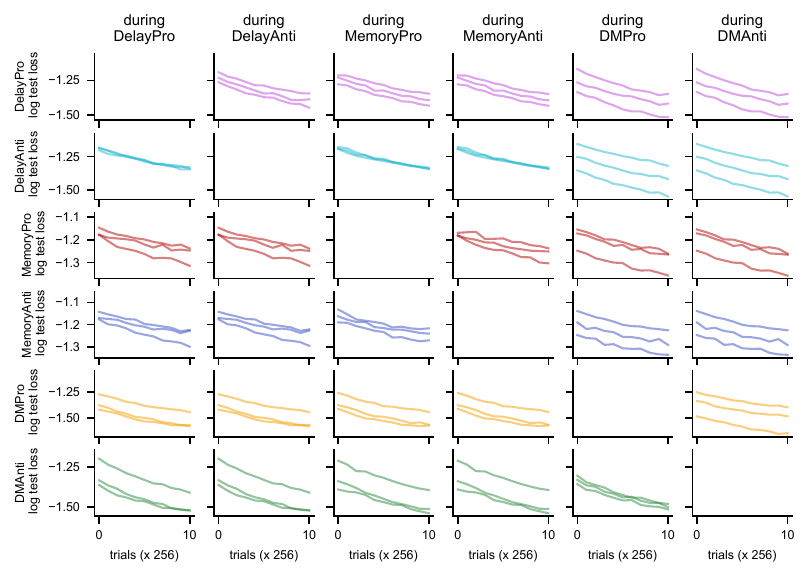}
% \vspace{-2ex}
\caption{Additional results on backward transfer. The log test loss of the previously trained task A (indicated by the row label) continue to decrease during subsequent training on another task B (column label). Each task is trained with 2560 trials and we plot results across three random seeds for each task order.}
\label{fig:backward_transfer_supp}
\end{figure}

\begin{figure}
\centering
\includegraphics{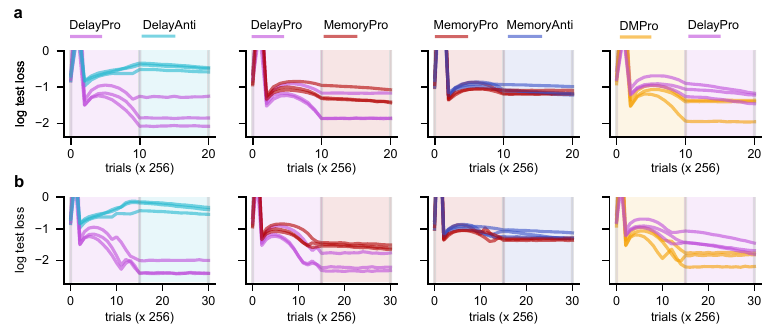}
% \vspace{-2ex}
\caption{No backward transfer with OWP. \textbf{a}: The log test loss during sequential training of two tasks with the OWP algorithm, each tasks is trained for 2560 trials. The loss of the previous tasks does not decrease after switching to a new task, indicating no backward transfer. Results are shown for three random seeds per task order. \textbf{b}: same as \textbf{a} but with each task trained for 3840 trials.}
\label{fig:backward_transfer_owp_supp}
\end{figure}

We provide additional results on transfer learning. Figure \ref{fig:forward_transfer_supp} and Figure \ref{fig:backward_transfer_supp} supplement Figure \ref{fig:transfer-learning}\textbf{a} and \textbf{b}, respectively, by showing results for all task pairs. Figure \ref{fig:backward_transfer_owp_supp} shows the lack of backward transfer when training with the OWP algorithm \cite{duncker2020organizing}, where the test loss of previously learned tasks did not decrease when training on subsequent tasks with overlapping epochs.

\subsection{Results with other hyperparameter choices}
\begin{figure}
\centering
\includegraphics{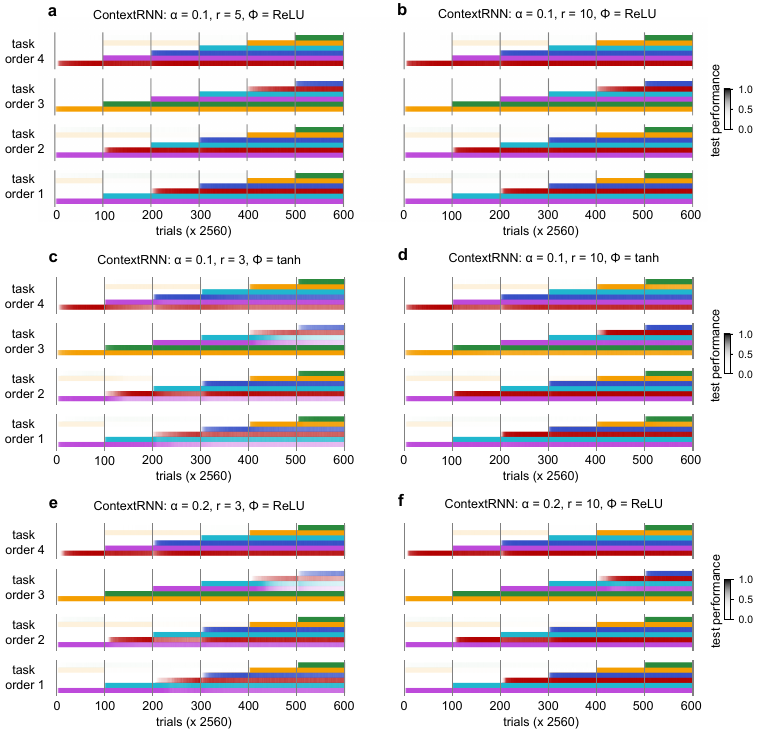}
% \vspace{-2ex}
\caption{Continual learning performance with different hyperparameter choices. \textbf{a}: Color-coded test performance during sequential training of four different task orders. Each row color-codes the average test performance across five random seeds of a specific task over training. RNNs used for this panel have $\alpha=0.1, r=5, \phi=\text{ReLU}$. \textbf{b-f}: same as \textbf{a} but with a different hyperparameter choice labeled by the title.}
\label{fig:rnn_hp_supp}
\end{figure}

We observed improved performance of our continual learning algorithm as we increased the rank of $U_z$ and $V_z$ (denoted by $r$) (Figure \ref{fig:rnn_hp_supp}\textbf{a},\textbf{b}, compared with Figure \ref{fig:rnn-continual-learning}\textbf{f}). Using $r = 3$, $5$, and $10$ in ContextRNN results in 34587, 43803, and 66843 trainable parameters, respectively --- all fewer than the 69379 parameters of a general RNN with the same number of hidden units. With $r=3$, ContextRNN performed worse with tanh than with ReLU activation, but this gap was closed at $r=10$ (Figure \ref{fig:rnn_hp_supp}\textbf{c}, \textbf{d}). Similarly, performance with $r=3$ and $\alpha=0.2$ was worse than with $\alpha=0.1$, but this gap was also closed at $r=10$ (Figure \ref{fig:rnn_hp_supp}\textbf{e}, \textbf{f}).

\section{Details of baseline implementations}
\label{appendix: other methods}

\subsection{Elastic Weight Consolidation}

Elastic Weight Consolidation (EWC) \cite{kirkpatrick2017overcoming}, slows down learning rates for network weights deemed important for previous tasks. This is achieved by adding a regularization term to the training loss function. For a loss $\mathcal{L}(\Theta)$, the EWC objective is given as
\begin{equation}
	\mathcal{L}^{EWC}(\Theta) = \mathcal{L}(\Theta) + \frac{\lambda}{2}\sum_i F_i (\Theta_i - \Theta_{i}^*)^2
\end{equation}
Here $F_i$ is an importance weight that ties the $i$th parameter value $\Theta_i$ to it's value $\Theta_i^*$ at the end of training on the previous task. The important weights are computed as the diagonal of the Fisher Information matrix $F$ evaluated at the parameter values $\Theta_{i}^*$. 
We set $\lambda$ to $10^5$ after a coarse grid search.

\subsection{Orthogonal Weight Projection}

% We compare our approach to a modified version of the continual learning rule \citet{duncker2020organizing} introduced for a similar set of tasks. 

The original learning rule in \citet{duncker2020organizing} was derived for a parameterization of network dynamics of the form
\begin{equation}
\dot{\vec x} =  -\vec x +  \phi\left( \vec W^{\mathsf{rec}} \vec x + \vec W^{\mathsf{in}} \vec s \right)
\label{eqn:rnn-nonlinear-linear}
\end{equation}
given an element-wise activation function $\phi(\cdot)$. The learning rule was intended to maintain the stimulus/response relationship of previous task by applying a set of projection matrices to the gradient used to update the network weights over learning. The projection matrices were intended to remove directions from the weight update that would interfere with previous tasks and were defined as follows. Letting $\vec z_t^{c,r} = \begin{bmatrix}
 \vec x_t^{c,r}\\ \vec s_t^{c,r}	
 \end{bmatrix}$ denote the concatenated network state and input state at time $t$ of trial $r$ on task $c$, $\vec Z_{1:c} = [\vec z_1^{1,1}, \dots \vec z_T^{c,r}]$ the collection of all time points and trials on tasks 1 through $c$, and $\vec W = [\vec W^{\mathsf{rec}} \;  \vec W^{\mathsf{in}}]$ the concatenated weight matrices, we define the projection matrices
 \begin{align}
 \vec P^{1:c}_1 & = \left(\frac{1}{\lambda} \vec Z_{1:c} \vec Z_{1:c} \tr + I \right) \inv\\
 \vec P^{1:c}_{2} & = \left( \frac{1}{\lambda} \vec W\vec Z_{1:c} \vec Z_{1:c} \tr  \vec W \tr + I\right) \inv
 \end{align}
and the modified learning update as
\begin{equation}
\Delta W^{\mathsf{CL}} \propto 	  \vec P^{1:c}_{2} \; \nabla_{\vec W} \mathcal{L} \; \vec P^{1:c}_1 
\end{equation}
where $\nabla_{\vec W} \mathcal{L}$ is the derivative of the loss on the new task with respect to the network weights. An analogous set of projection matrices and modified learning update is used for the readout weights.

% This approach was largely motivated by intuition from the linear setting. Consequently, the authors showed that this worked best for $\phi(\cdot) = \max(\cdot, 0)$, but that the continual learning performance degraded for more general choices of nonlinearity, which we unpack further after introducing our modification below.

To facilitate direct comparisons with our approach, we adapted the learning rule to a modified setting, where the RNN dynamics are expressed as
\begin{equation}
\dot{\vec h}= - \vec h +  \vec W^{\mathsf{rec}} \phi (\vec h) + \vec W^{\mathsf{in}} \vec s
\label{eqn:rnn-linear-nonlinear}
\end{equation}
given the same element-wise activation function $\phi(\cdot)$. While the two parameterizations in (\ref{eqn:rnn-nonlinear-linear}) and (\ref{eqn:rnn-linear-nonlinear}) are generally considered equivalent \citep{miller2012mathematical}, the linear intuition used to motivate the approach of \cite{duncker2020organizing} should be exact in (\ref{eqn:rnn-linear-nonlinear}). With $\vec x_t = \phi(\vec h_t)$, $\vec Z_{1:c}$ is unchanged, but we instead use
\begin{align}
 \vec P^{1:c}_{2} & = \left( \frac{1}{\lambda} \vec H_{1:c} \vec H_{1:c} \tr  + I\right) \inv
 \end{align}	
where  $\vec H_{1:c} = [\vec h_1^{1,1}, \dots \vec h_T^{c,r}]$. While this is very similar to the version in \citet{duncker2020organizing}, the projections matrix now only depends implicitly on the $\vec W$ of previous tasks. We performed all comparisons using this modified learning rule.

\subsection{Neuromodulated RNN}
\citet{costacurta2024structured} designed a neuromodulated RNN (NM-RNN), in which a neuromodulatory subnetwork (RNN1) outputs a low-dimensional signal that dynamically scales the low-rank recurrent weights of an output-generating RNN (RNN2). RNN1 and RNN2 are jointly optimized through gradient descent. In our implementation, RNN2 has 256 neurons and with rank(${\mathbf W^{rec}}$) = 27, equal to the default choice for our ContextRNN. Number of neurons in RNN1 is set to 125 such that the whole network has 36658 parameters, comparable to the 34587 parameters of our contextRNN. $\alpha$ is set to 0.1 for RNN2 and 0.01 for RNN1 following the original paper. We used Adam optimizer, with learning rate set to 0.001 and $L_2$ weight regularization coefficient set to $10^{-5}$ after a coarse grid search.

\subsection{Hypernetwork}
\citet{von2019continual} designed task-conditioned hypernetworks to tackle continual learning. A hypernetwork receives distinct learnable embeddings for each task, and outputs the weights for the target network (an output-generating RNN in our case). For the output-generating RNN, we set the number of neurons to 256. A full-rank recurrent weight matrix, an input weight matrix and an output weight matrix together require a 67843-d output from the hypernetwork. We used a chunked hypernetwork introduced in the original paper, where a set of learnable chunk embeddings serve as additional input to the hypernetwork. Concatenating the hypernetwork output from distinct chunks gives all parameters for the target network. We set the output dimension of the hypernetwork to 2000 and number of chunks to 34. Dimension of task embeddings and chunk embeddings is set to 32. The hypernetwork is a multi-layer perceptron with two 32-d hidden layers. Together, the number of trainable parameters of the hypernetwork equals 70416, comparable to other baselines in our paper using a full-rank RNN. After a coarse grid search, we set continual learning regularization strength to 1, orthogonal regularization strength to 0.001 and clip gradient norm to 1 (see \citet{ehret2020continual} for details). We used Adam optimizer, with learning rate set to 0.001 and $L_2$ weight regularization coefficient set to $10^{-5}$.
% \section{Technical Appendices and Supplementary Material}
% Technical appendices with additional results, figures, graphs and proofs may be submitted with the paper submission before the full submission deadline (see above), or as a separate PDF in the ZIP file below before the supplementary material deadline. There is no page limit for the technical appendices.

%%%%%%%%%%%%%%%%%%%%%%%%%%%%%%%%%%%%%%%%%%%%%%%%%%%%%%%%%%%%

\end{document}